\documentclass{article}
\pdfoutput=1

\usepackage{wrapfig}

\usepackage{xcolor} 
\definecolor{boxcolor}{RGB}{51,51,153}
\definecolor{fg}{HTML}{009B55}
\usepackage{colortbl}
\usepackage{graphicx}
\usepackage{times}
\usepackage{latexsym}

\usepackage[T1]{fontenc}

\usepackage[utf8]{inputenc}

\usepackage{microtype}

\usepackage{inconsolata}
\usepackage[inkscapelatex=false]{svg}
\usepackage{latexsym}
\usepackage{multirow}
\usepackage{booktabs}
\usepackage[T1]{fontenc}
\usepackage{soul}
\usepackage{CJKutf8} 
\usepackage{pdflscape}
\usepackage[utf8]{inputenc}
\usepackage{threeparttable}
\usepackage{tabularx}
\usepackage{microtype}
\usepackage[figuresright]{rotating}
\usepackage{inconsolata}
\usepackage{xcolor}
\usepackage{amssymb}
\usepackage{pifont}
\usepackage{listings}

\usepackage{longtable}
\usepackage{placeins}
\usepackage[most]{tcolorbox}
\usepackage{float}
\usepackage{xspace}
\tcbset{
  aibox/.style={
    width=474.18663pt,
    top=10pt,
    colback=white,
    colframe=black,
    colbacktitle=black,
    enhanced,
    center,
    attach boxed title to top left={yshift=-0.1in,xshift=0.15in},
    boxed title style={boxrule=0pt,colframe=white,},
  }
}
\newtcolorbox{AIbox}[2][]{aibox,title=#2,#1}

\usepackage[hidelinks]{hyperref}  

\usepackage[preprint]{neurips_data_2024}

\usepackage{array}

\newcolumntype{P}[1]{>{\centering\arraybackslash}p{#1}} 
\definecolor{darkgreen}{rgb}{0.0, 0.5, 0.0}
\definecolor{maroon}{rgb}{0.5, 0.0, 0.0}
\definecolor{navy}{rgb}{0.0, 0.0, 0.5} 
\definecolor{teal}{rgb}{0.0, 0.5, 0.5} 

\definecolor{llavacolor}{rgb}{0.0, 0.5, 0.0}
\definecolor{lviscolor}{rgb}{0.5, 0.0, 0.0}

\usepackage[utf8]{inputenc} 
\usepackage[T1]{fontenc}    
\usepackage{hyperref}       
\usepackage{url}            
\usepackage{booktabs}       
\usepackage{amsfonts}       
\usepackage{nicefrac}       
\usepackage{microtype}      
\usepackage{xcolor}         

\title{MLLM-Bench: Evaluating Multimodal LLMs\\with Per-sample Criteria}

\author{%
Wentao Ge$^*$, 
Shunian Chen$^*$, 
Guiming Hardy Chen\thanks{The first three authors contribute to this work equally.}~,
\textbf{Junying Chen}, 
Zhihong Chen$^\dagger$, \\
\textbf{Nuo Chen},
\textbf{Wenya Xie}, 
\textbf{Shuo Yan}, 
\textbf{Chenghao Zhu}, 
\textbf{Ziyue Lin}, 
\textbf{Dingjie Song}, \\
\textbf{Xidong Wang}, 
\textbf{Anningzhe Gao}, 
\textbf{Zhiyi Zhang}, 
\textbf{Jianquan Li}, 
\textbf{Xiang Wan}, \\
\textbf{Benyou Wang}\thanks{%
Zhihong and Benyou are the corresponding authors.}\\
Shenzhen Research Institute of Big Data\\
The Chinese University of Hong Kong, Shenzhen\\
\texttt{zhihongchen@link.cuhk.edu.cn}, \texttt{wangbenyou@cuhk.edu.cn} \\
}

\begin{document}

\maketitle

\begin{abstract}
Multimodal large language models (MLLMs)  have broadened the scope of AI applications. 
Existing automatic evaluation methodologies for MLLMs are mainly limited in evaluating \textit{objective} queries without considering \textit{real-world} user experiences, inadequately addressing the nuances of creative and associative multimodal tasks. However, the \textit{open-ended} and \textit{subjective} nature of such tasks poses a significant challenge to the evaluation methodology, where it is difficult to define the ground-truth answers for them.
To this end, in our paper, we propose a new evaluation paradigm for MLLMs, which is evaluating MLLMs with \textit{per-sample criteria} using potent MLLM as the judge. 
To validate the feasibility and effectiveness of this paradigm, we design a benchmark, dubbed MLLM-Bench, by curating the evaluation samples across six comprehensive cognitive levels.
We benchmark 21 popular MLLMs in a pairwise-comparison fashion, showing diverse performance across models.
Moreover, the validity of our benchmark manifests itself in reaching 88.02\% agreement with human evaluation.
We contend that the proposed paradigm explores the potential of MLLMs as effective evaluation tools with the help of per-sample criteria.
See online leaderboard at \url{https://mllm-bench.llmzoo.com}.
\end{abstract}


\section{Introduction}



\begin{table*}[ht]
\footnotesize
\caption{\label{tab:exmp_pair_vote} \small Pair-wise evaluation using per-sample criteria for MLLMs. The per-sample criteria is only available to the GPT-4V judge and are not accessible to evaluated MLLMs such as {\color{llavacolor} LLaVA-v1.5-13B} and {\color{lviscolor} LVIS-instruct4v-LLaVA-7B.}}
\vspace{1em}
\centering
{\renewcommand{\arraystretch}{0.6}
\resizebox{\textwidth}{!}{
\begin{tabular}{p{0.14\textwidth}p{0.85\textwidth}} 
\toprule
\vspace{-6pt}
\multirow{2}{*}{\raisebox{1\height}{\includegraphics[height=1.10in]{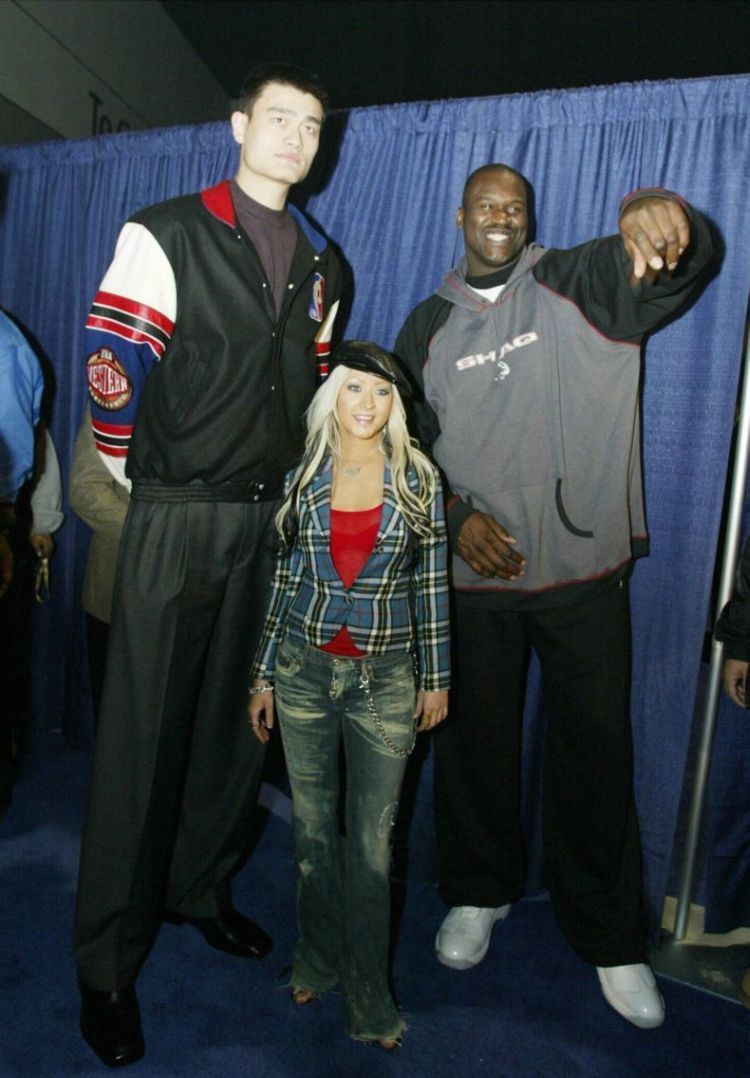}}} &
\begin{tabular}{p{0.9\textwidth}}
~~~~~\textbf{Question}: Based on the visual cues, estimate the approximate height of the woman.\\
\end{tabular} \\ [5pt] 
\cline{2-2} \\  [.5pt] 
& \begin{tabular}{p{.35\textwidth}p{.55\textwidth}} 
\textbf{{\color{llavacolor}Answer 1 ~(LLaVA-v1.5-13B):}} & \textbf{{\color{lviscolor} Answer 2~(LVIS-instruct4v-LLaVA-7B):}} \\
The woman in the image is relatively short, standing at around 3 feet tall. &  
The woman is standing between two tall men, and based on the visual cues, she appears to be quite short in comparison. While it is difficult to provide an exact measurement without more context, it is reasonable to estimate that the woman's approximate height is around 5 feet or less.\\
\end{tabular} \\
\vspace{5pt}
\\
\hline
\\
\rowcolor{gray!25}
\multicolumn{2}{p{1.12\textwidth}}{\textbf{Criteria}} \\
\rowcolor{gray!25}
\multicolumn{2}{p{1.12\textwidth}}{ \textit{1. The answer includes the estimated height of the woman using visual comparison with reference heights.}} \\
\rowcolor{gray!25}
\multicolumn{2}{p{1.12\textwidth}}{ \textit{2. Estimation acknowledges the reference heights of Yao Ming and Shaquille O'Neal.}} \\
\rowcolor{gray!25}
\multicolumn{2}{p{1.12\textwidth}}{\textit{3. The height range of the woman should be 165cm to 175cm.}} \\
\midrule
\\[.1pt]
\multicolumn{2}{c}{\textbf{GPT-4V Judgment}} \\
\multicolumn{2}{p{1.12\textwidth}}{
Upon reviewing the provided responses in relation to the visual cues from the image and the given evaluation criteria:

{\color{llavacolor} Answer 1} suggests that the woman is ``around 3 feet tall.'' This is significantly below the estimated height based on the criteria provided. Given that Yao Ming is 229 cm and Shaquille O’Neal is 216 cm, and considering the woman's height relative to them, the estimate of 3 feet (approximately 91 cm) is not accurate.

{\color{lviscolor} Answer 2} estimates the woman's height to be ``around 5 feet or less''. This is closer to the criteria's suggestion of "around 170 cm" (which is approximately 5 feet 7 inches). While "5 feet or less" is a broad range and could include the correct height, the answer does not commit to inaccuracy as specific as {\color{llavacolor} Answer 1}.

Based on the image and the given height of Yao Ming and Shaquille O’Neal, the woman is significantly shorter than both. However, she is not as short as 3 feet, as suggested by {\color{llavacolor} Answer 1}. {\color{lviscolor} Answer 2}, while not giving a precise height, provides a range that is more plausible and closer to the criteria's approximation.

Therefore, {\color{lviscolor} Answer 2} is the superior response based on the quality of the answer, its alignment with the image, and the provided evaluation criteria.
\vspace{-1pt}
}\\
\bottomrule
\end{tabular}
}
}
\end{table*}



The field of artificial intelligence has witnessed a paradigm shift with the incorporation of multimodal inputs, particularly the amalgamation of visual and linguistic data, mirroring the complex processing capabilities of the human brain. The development of multimodal large language models (MLLMs) (e.g., MiniGPT-4~\citep{zhu2023minigpt}, LLaVA~\citep{liu2023improved,liu2023llava}, Qwen-VL~\citep{bai2023qwenvl}, and GPT-4V~\citep{GPT4V}) represents a leap towards more sophisticated, context-aware AI systems. These models are increasingly crucial for tasks that demand an understanding of both visual cues and textual content. However, the expansion of capabilities brings forth the challenge of evaluation – how does one accurately measure the effectiveness of a system designed to mimic the inherently subjective and associative processes of human perception?


The predominant evaluation frameworks for MLLMs focus on close-ended queries with clear-cut, correct answers. Such tasks are valuable in quantitating the model performance but do not consider user experience and encapsulate the full spectrum of human-like cognitive tasks that modern MLLMs aim to perform. The realms of creativity, association, and ethical judgment within multimodal contexts, for instance, resist reduction to simple right or wrong answers. However, human evaluation is very costly and inefficient, while automatically evaluating the MLLMs' performance on open-ended queries is challenging. This limitation in evaluation methodologies leads to an incomplete understanding of a model's capabilities and fails to provide insight into how such models might interact with users in real-world scenarios, where answers are often nuanced and context-dependent.


To bridge this gap, we propose to use potent MLLM~\footnote{ We adopt GPT-4V as  judges for main experiments, while we also shows  result using Cluade-3-Opus~\citep{Claude3}.} as the judge with \textbf{per-sample criteria} to evaluate MLLMs. To validate this paradigm, we develop a comprehensive benchmarking suite, named MLLM-Bench, including 42 distinct aspects of MLLM functionality, distributed among six critical levels of capability: \textit{perception}, \textit{understanding}, \textit{applying}, \textit{analyzing}, \textit{evaluating}, and \textit{creation}, inspired by Bloom's Taxonomy~\citep{krathwohl2002revision}. 
Notably, rather than providing a standard answer to a posed question, we offer one to three specific \textit{evaluation criteria for each question}. These criteria are then applied in conjunction with the posed question and the model's response to assess performance more accurately and contextually (see an example in Figure~\ref{tab:exmp_pair_vote}\footnote{The GPT-4V judgment  is for demonstration purposes which might be  different from our experiment.}).

The contributions of this paper are two-fold:
1) We propose \textbf{a new  paradigm to evaluate MLLMs} utilizing powerful MLLMs along with per-sample criteria,  which shifts from traditional, fixed-answer evaluations for MLLMs to a flexible, criteria-based approach, particularly suited for open-ended tasks.   
It recognizes and acknowledges a spectrum of valid responses and evaluates the answer quality based on how well they align with these criteria, assessing models beyond the single ``correct'' answer limitation. 
2) A benchmark \textbf{dataset} with a comprehensive taxonomy that categorizes tasks and scenarios for MLLMs, with a strong emphasis on ethical considerations. In the benchmark, we conduct a systematic \textbf{benchmarking} of existing MLLMs; the benchmarking is unique since it might be well-aligned to user experience in real-world applications where users usually raise questions without static standard answers.






\section{Background: Evaluations for MLLMs}


It is challenging to comprehensively  assess the  capabilities of MLLMs . Current benchmarks primarily fall into several categories: (1) Multiple-choice questions (evaluating the perception and cognition abilities of MLLMs): MME~\citep{fu2023mme}, SEED~\citep{li2023seed}, and TouchStone~\citep{bai2023touchstone}; (2) Arena-like evaluation (user-based evaluation of different capabilities): LVLM-eHub~\citep{xu2023lvlmhub}, VisIT-Bench~\citep{bitton2023visitbench}; (3) Hallucination assessment (focusing on a key issue currently faced by MLLMs - hallucinations): POPE~\citep{li2023rope} and HallusionBench~\citep{liu2023hallusionbench}.
The works most related to us are (i) MMBench~\citep{liu2023mmbench} and MM-Vet~\citep{yu2023mmvet}, using GPT-4 as the evaluator to quantitatively measure the performance of different MLLMs; (ii) a concurrent work~\citep{zhang2023gpt}  uses GPT-4V to evaluate  text-to-image generation.

\paragraph{Open and Closed-ended Benchmarks}

Existing benchmarks are categorized into either open-ended or close-ended, reflecting different evaluation approaches.
Although close-ended benchmarks such as MMLU~\citep{hendrycks2020measuring}, C-Eval~\citep{huang2023ceval} for LLMs and  MME~\citep{fu2023mme} and SEED~\citep{li2023seed} for MLLMs  are convenient to evaluate on, they often suffer from data contamination issue. The results of close-ended benchmarks are especially for proprietary LLMs whose training data are all \textit{in-house}.
On the other hand, open-ended benchmarks (e.g., MT-Bench~\citep{zheng2023judging} and Alpaca-Eval~\citep{alpaca_eval}) test models via free-form generation, which is more consistent with real-world use cases and relies heavily on LLMs' generation ability. The issue of data contamination in open-ended benchmarks is less severe since there are no standard answers, and such contamination offers minimal assistance in benchmark hacking.


\paragraph{LLMs for MLLM Evaluation }
MMBench~\citep{liu2023mmbench}, TouchStone~\citep{bai2023touchstone}, and MM-Vet~\citep{yu2023mmvet} employ LLM-based evaluation frameworks, leveraging the capabilities of advanced LLMs (\textit{e.g.,} GPT-4) for assessing MLLMs. This approach encounters significant limitations due to the inherent inability of pure language models to perceive visual contexts directly. 
The idea of adopting GPT-4V, a potent MLLM, directly as a judge in this paper, marks a significant advancement in the field.

\section{Motivation of MLLM Evaluation with Per-sample Criteria}

\subsection{Motivations}
\paragraph{Why MLLM-as-the-judge evaluation needs additional criteria?}
While potent MLLMs are potential evaluators, their assessment outcomes may not always align perfectly with factual accuracy or human standards. This discrepancy highlights the necessity for a more nuanced approach to evaluation: per-sample criteria. Per-sample criteria are designed to provide specific benchmarks and guidelines for each assessment task, aiding MLLM judge for MLLM evaluation. This approach is particularly valuable for tasks where the judge's capabilities might fall short, ensuring that evaluations remain robust even in areas of potential weakness. Unlike a single reference answer, per-sample criteria afford a broader and more flexible basis for assessment, making them ideally suited for evaluating open-ended questions. 
As we adopt GPT-4V as the judge in this paper, an example where GPT-4V falls short is shown in App.~\ref{sec:4v_eval}.


\paragraph{Why criteria should be sample-specific?}
Different samples present unique challenges and requirements, making a one-size-fits-all approach to evaluation inadequate. 
Table~\ref{tab:typical_per_sample_criteria} exemplifies the critical per-sample criteria essential for the nuanced evaluation of multi-modal large language models (MLLMs), especially in \textit{Soft Reference}, \textit{Range}, and  \textit{Evaluation Guideline} catogories. 
These criteria underscore the need for evaluations that adapt to the task's context and intricacies, showcasing the limitations of previous methodologies in comprehensively assessing MLLM capabilities.

\begin{table}[ht]
\centering
\footnotesize
\caption{Typical per-sample criteria.  The criteria are sampled from this benchmark.}
\setlength{\tabcolsep}{3pt} 
\begin{tabular}{p{0.15\columnwidth}p{0.4\columnwidth}p{0.4\columnwidth}}
\toprule
\textbf{Criteria Type} & \textbf{Description} & \textbf{Example criteria (images omitted)} \\
\midrule
{Exact Reference} & Necessitates precise answers, such as accurately determining an item's location. & \textit{Answer specifies cactus location as row 2, column 6.}\\  \midrule
{Soft Reference} & Requires identifying and translating contextually relevant texts, demanding linguistic adeptness and current knowledge. & \textit{Identify and translate text related to the novel coronavirus pneumonia in Wuhan.} \\
 \midrule
{Range} & Allows for variability within defined limits, like estimating a person's height, introducing flexibility. & \textit{ The height range of the woman should be 165cm to 175cm.} \\
 \midrule
{Evaluation Guideline} & Involves interpretative analysis to evaluate complex situational effects on responses. & \textit{ Assess if the answer considers the environment's condition, like submerged paths.} \\
\bottomrule
\end{tabular}
\label{tab:typical_per_sample_criteria}
\end{table}



\subsection{Benefits of Per-sample Criteria }

\paragraph{Generalization of Referenced-based Evaluation}
For questions with \textit{objective} answers, the criteria offer specific reference answers, as demonstrated in the first, third, and fifth samples in Table~\ref{tab:sample}. 
In cases where the instructions require a \textit{subjective} description or yield ambiguous answers, the criteria supply essential information for formulating responses, exemplified by the second,fourth, and sixth samples in the same table. 
This approach enables the evaluation model (GPT-4V) to assess the quality of outputs using a well-defined standard.


\paragraph{Mitigation of Data Contamination} 
One of the advantages of our per-sample criteria is its potential to alleviate data contamination problems. While we will continuously update our dataset, there is no guarantee that the samples have not been seen or used. To this end, we choose to withhold the per-sample criteria when publishing our dataset. Namely, the released version will only include the images and instructions. 
We believe that this approach significantly reduces the risk of contamination since even if models have been exposed to the images, they do not necessarily generate desired responses.






\section{MLLM-Bench Dataset}

\label{sec:bench}


\subsection{Taxonomy of  Capabilities}
Due to the absence of a standardized framework for categorizing the capabilities of multimodal large language models, and acknowledging that vision-language models emulate human cognitive processes to a certain extent, we have chosen to adopt the revised Bloom's Taxonomy \citep{krathwohl2002revision} as the framework for this benchmark. In reference to Bloom's Taxonomy, we manually conclude 42 capabilities of MLLMs across a hierarchy spanning six cognitive levels and create 10 questions for each capability.  The six capability levels are shown below in Table~\ref{table:stats_and_desc}. For each of the capabilities, we create 10 questions, resulting in a total of 420 image-instruction pairs, see details in App.~\ref{sec:taxonomy}.

\subsection{Data Annotation}

\subsubsection{Data Annotators}
We have recruited six volunteers, all of whom are undergraduates, graduate students, or research assistants at a university with an all-English curriculum. Each volunteer is paid according to the local salary (i.e., equivalent to roughly 10 dollars per hour) and tasked with gathering data pertinent to a distinct capability level, thereby guaranteeing consistency within each specific category. The data collection phase spanned a duration of two weeks; they can complete it whenever convenient. Before data collection and annotation, they are instructed to follow a guideline. See details in App.~\ref{sec:task_distribution}.


\begin{table}[t]
\centering
\footnotesize
\caption{Taxonomy of MLLM-Bench including examples  at six cognitive levels.} 
\setlength{\tabcolsep}{3pt} 
\begin{tabular}{m{0.15\columnwidth}cp{0.7\columnwidth}}
\toprule
\textbf{Level} & \textbf{\#Samples} & \textbf{Description} \\
\midrule
Perception & 70 & MLLMs retrieve information from multimodal inputs, using skills like object recognition and OCR. \\ \midrule
Understanding & 110 & MLLMs process perceived information to construct meaning, comprehending and interpreting data contextually. \\ \midrule
Applying & 60 & MLLMs apply knowledge to similar situations, such as using text-based knowledge to interpret images as in medical imaging.\\ \midrule
Analyzing & 120 & MLLMs break down information to explore relationships, performing tasks like attribute comparison or causal reasoning. \\ \midrule
Evaluation & 40 & MLLMs make judgments based on criteria and standards, like assessing image quality or discerning content authenticity.\\ \midrule
Creation & 20 & MLLMs synthesize information to generate new content, from visual storytelling to coding with vision. \\ 
\bottomrule
\end{tabular}
\label{table:stats_and_desc}
\end{table}

\vspace{-1.7mm}
\subsubsection{Pipeline}
\label{sec:data_ppl}

The data collection and annotation module comprises four stages:

\textbf{I: Image Collection}: Volunteers gather the most recent images that are pertinent to the capabilities under examination, either from social networks or by capturing them in real life.

\textbf{II: Instruction Construction}: Utilizing GPT-4V, volunteers craft assessment instructions that are congruent with the requisite model capabilities, the context of the collected images, and the manually written prompts. This stage also includes a consistency check between the instructions and the corresponding images.

\textbf{III: Question Type Annotation}: Recognizing that our benchmark evaluates model performance on both traditional closed-ended questions and more exploratory open-ended tasks, volunteers are required to annotate the type of each question. This includes categorizing them as \textit{open-ended}, \textit{closed-ended}, or \textit{compound}. Compound questions contain elements of both open-ended and closed-ended queries, offering a more comprehensive challenge to the models being tested.

\textbf{IV: Per-Sample Criteria Annotation}: Volunteers are asked to provide evaluation criteria based on the image, instructions, capability requirements, and question type for each item of data. These criteria include \textit{exact references}, \textit{soft references}, \textit{acceptable ranges}, and \textit{evaluation guidelines}.


\subsubsection{Guidelines for Annotators}

\textbf{Data Protocol}
As illustrated in Table~\ref{table:stats_and_desc}, each entry in the MLLM-Bench dataset comprises three key components:
1) a contemporary \textbf{image} with a friendly license,  2) a \textbf{question} posed as it would naturally arise in real-world situations and 3) 1-3 customized \textbf{per-sample criteria} that offer guidelines specifically designed to complement the capabilities of GPT-4V, thus enabling a more logical and scientifically sound evaluation.

\textbf{Guideline for the Data Annotation}
\label{sec:anno_guideline}
The guideline for data annotation  emphasizes the importance of using recent images to avoid data leakage\footnote{To mitigate the possible data leakage issue that collected data could be used as a part of training for evaluated models, one way is to continuously maintain and expand the existing dataset in real-time.}, sourcing data from publicly licensed platforms like Twitter or direct captures with clear copyright status, ensuring image clarity while accommodating real-world quality variance, maintaining impartiality by excluding sensitive content, and promoting diversity in response formats to reflect complex real-world interactions. See details in  \ref{sec:Guideline}. 

\begin{table*}[t]

\large
\centering
\caption{\label{tab:sample} Data samples in MLLM-Bench, which are presented from top to bottom across six capability levels: Perception, Understanding, Applying, Analyzing, Evaluation, and Creation.}
\vspace{1em}
\resizebox{\textwidth}{!}{
\begin{tabular}{>{\centering\arraybackslash}m{2cm} >{\centering\arraybackslash}m{3cm} >{\arraybackslash}m{8cm} >{\arraybackslash}m{8cm} }
\toprule
\textbf{Capability} & \textbf{Image} & \multicolumn{1}{c}{\textbf{Sample Questions}} & \multicolumn{1}{c}{\textbf{Criteria}}\\
\midrule
\begin{tabular}{@{}c@{}} Food \\ Recognition\end{tabular}  & 
\centering\arraybackslash\includegraphics[height=0.7in]{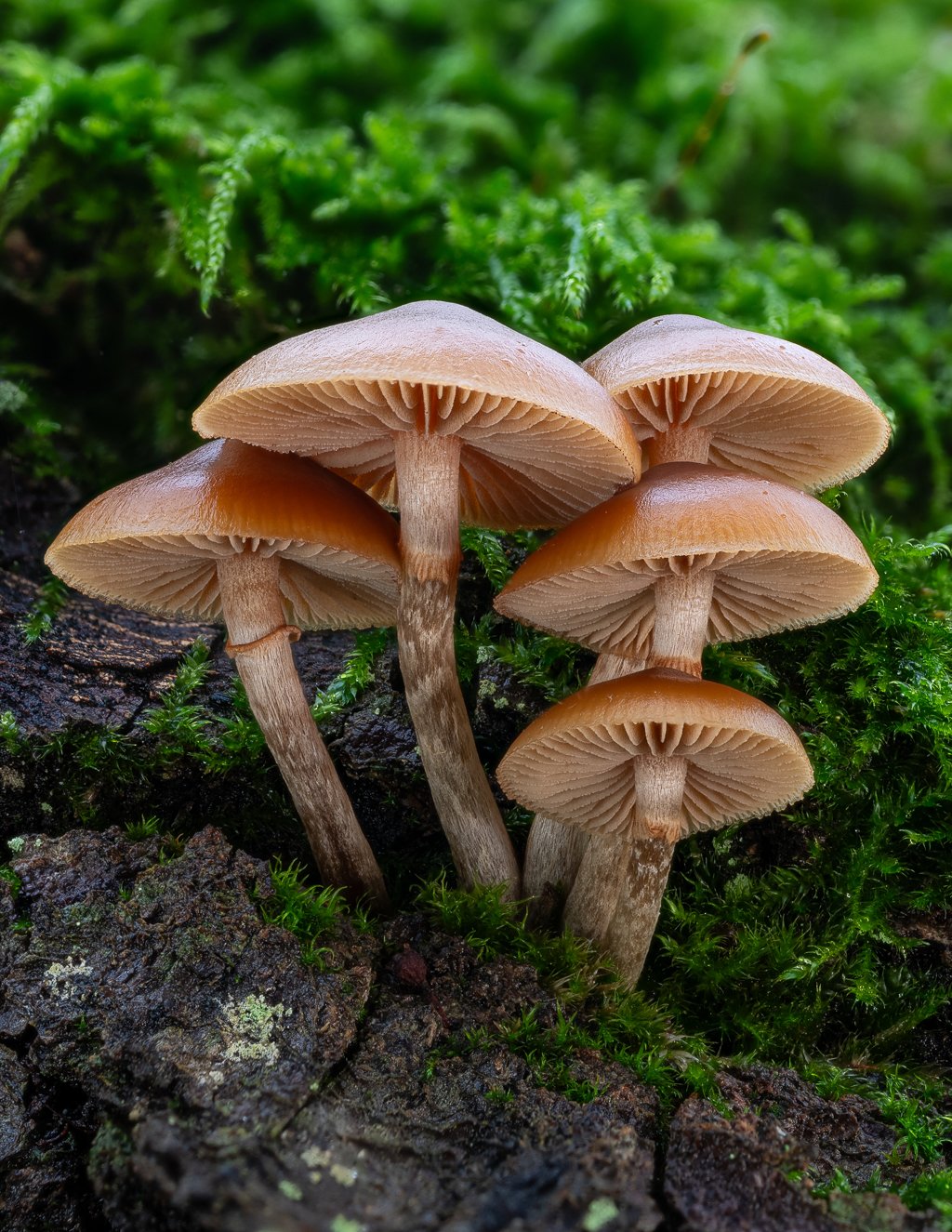}
& What type of mushroom is it in the image? Is it edible? & \textbf{1.} Identify mushroom as galerina marginata. \textbf{2.} State the mushroom is not edible. \textbf{3.} Ensure the answer is based on visual characteristics of the mushroom in the image. \\
\hline
\begin{tabular}{@{}c@{}} Attribute \\ Recognition\end{tabular}  & \centering\arraybackslash\includegraphics[height=.7in]{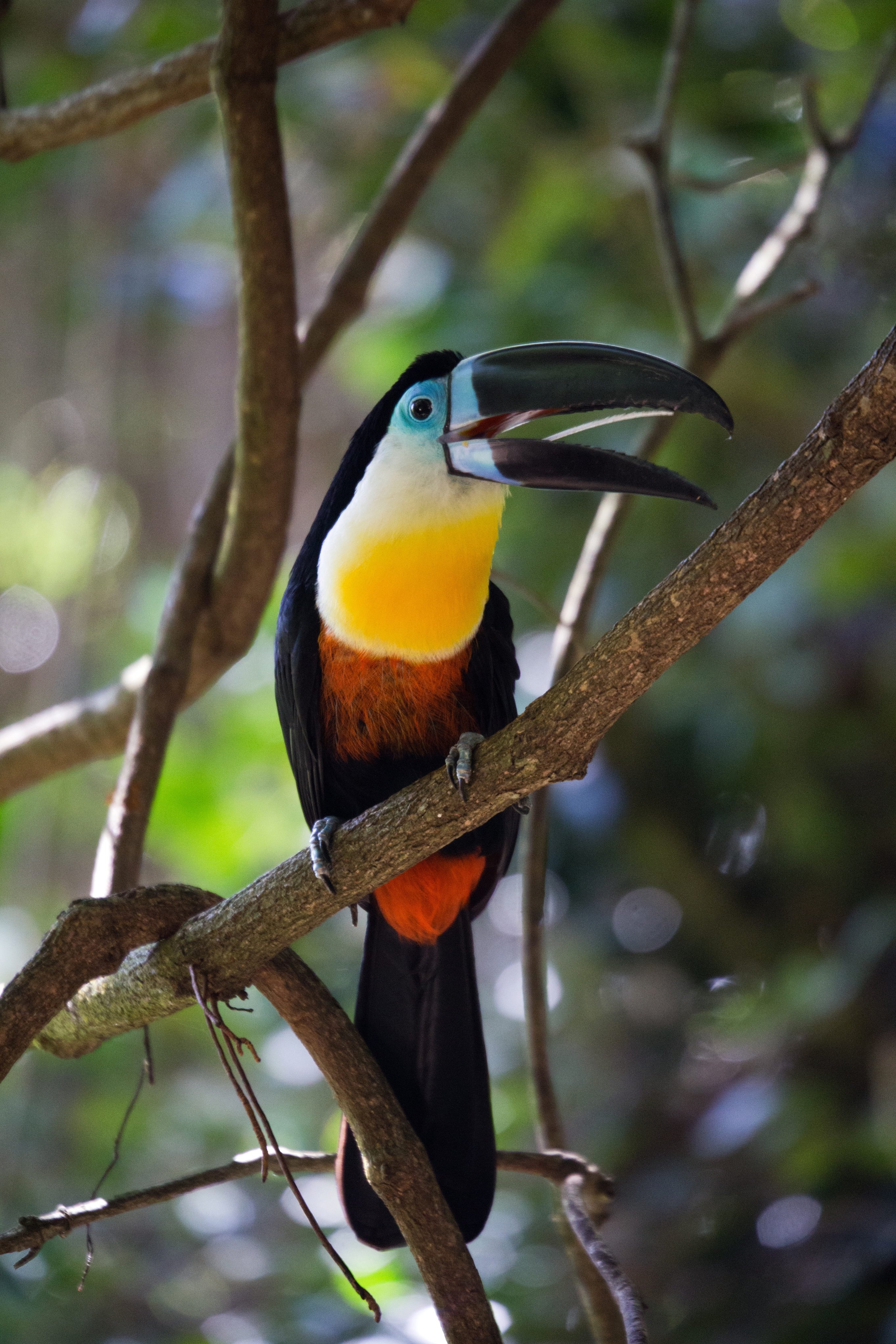} & Please describe the feature of the animal as detail as possible to help other people to recognize it. & \textbf{1.} Mention toucans' large, colorful bills as a distinctive feature. \textbf{2.} Describe additional attributes: body color, size, and any unique markings. \textbf{3.} Include the setting or behavior if it aids in recognition. \\
\hline
\begin{tabular}{@{}c@{}} Object \\ Localization\end{tabular}     & \centering\arraybackslash\includegraphics[height=0.7in]{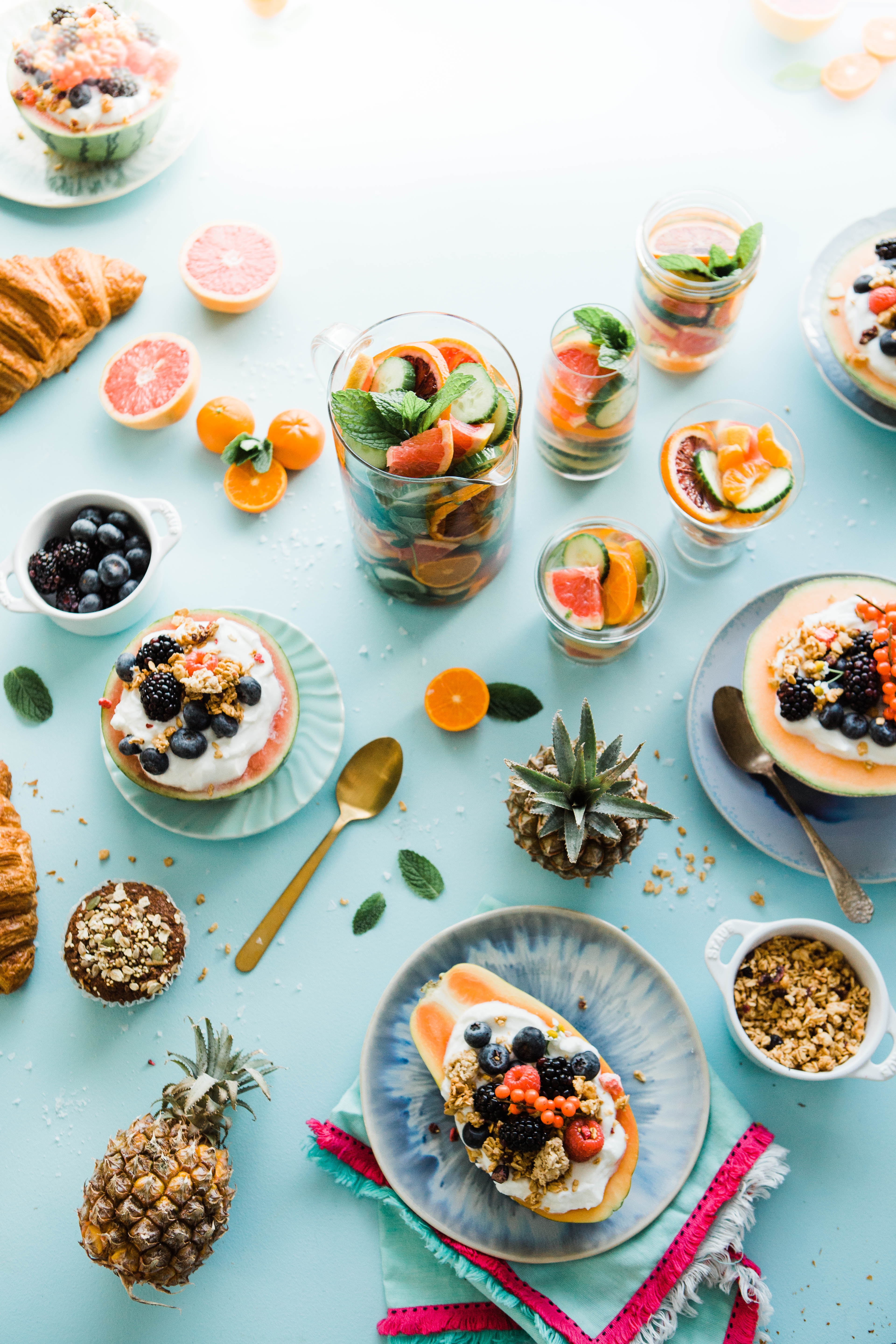} & The image is of dimension 700*1050, assume the bottom left corner to be the origin of coordinates (0,0), the coordinates of the top left corner to be (0,1050) and the coordinates of the bottom right corner to be (700,0). Please give the approximate coordinates of the spoon tail. & \textbf{1.} Answer should include coordinates close to the reference: (200,315), the closer the better. \\
\hline
\begin{tabular}{@{}c@{}} Function \\ Reasoning\end{tabular}   & \centering\arraybackslash\includegraphics[height=0.7in]{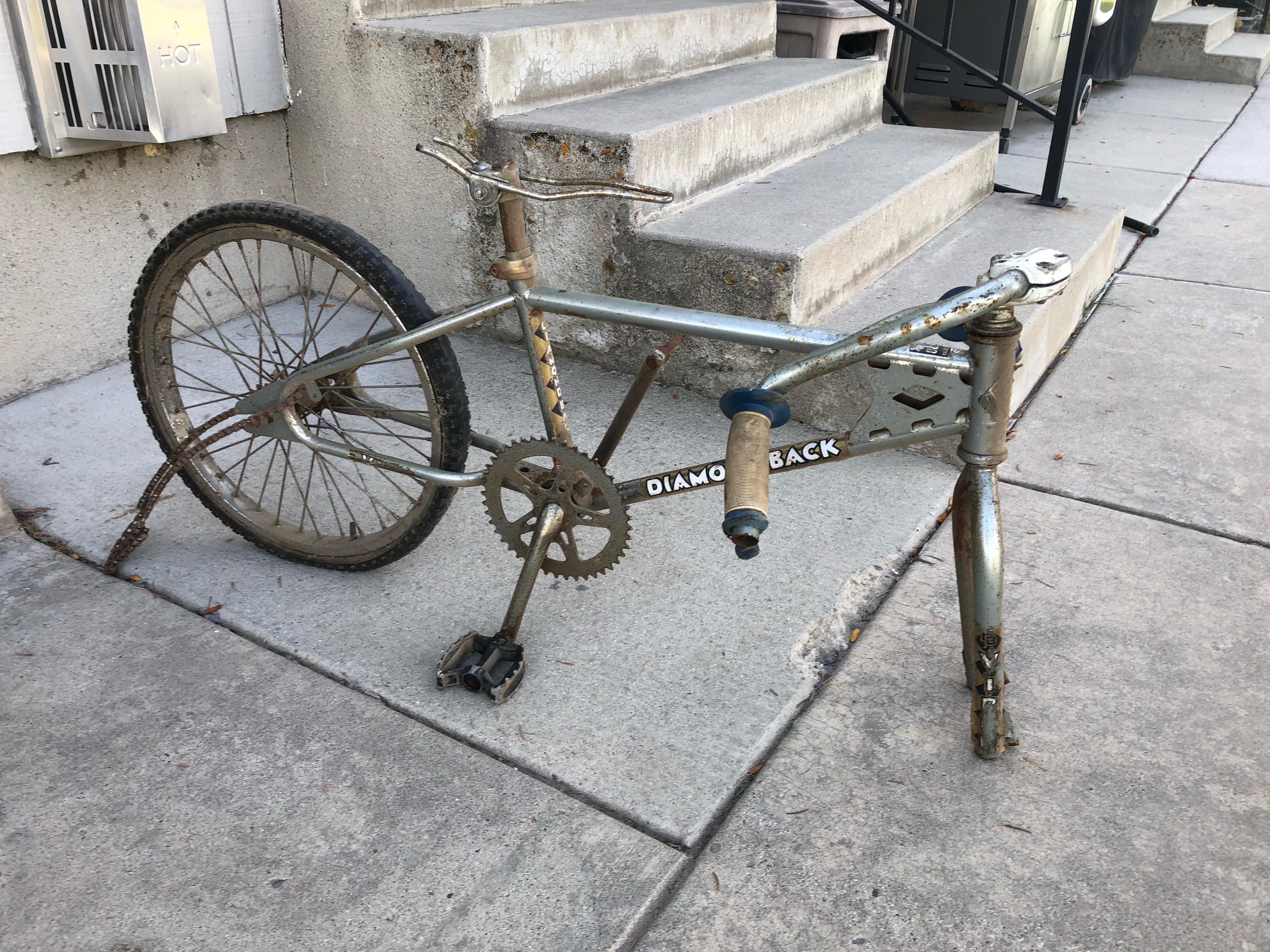} & The image depicts a bike that has been a part of my collection for quite some time. If I were to consider riding it, could you please advise me on the components required to restore it to working condition? & \textbf{1.} Identify missing/damaged components: front tire, chain, pedal, handle bar, seat. \textbf{2.} Explain the function of each identified component for bike restoration. \\
\hline
\begin{tabular}{@{}c@{}} Fake Image \\ Detection\end{tabular}   & \centering\arraybackslash\includegraphics[height=0.7in]{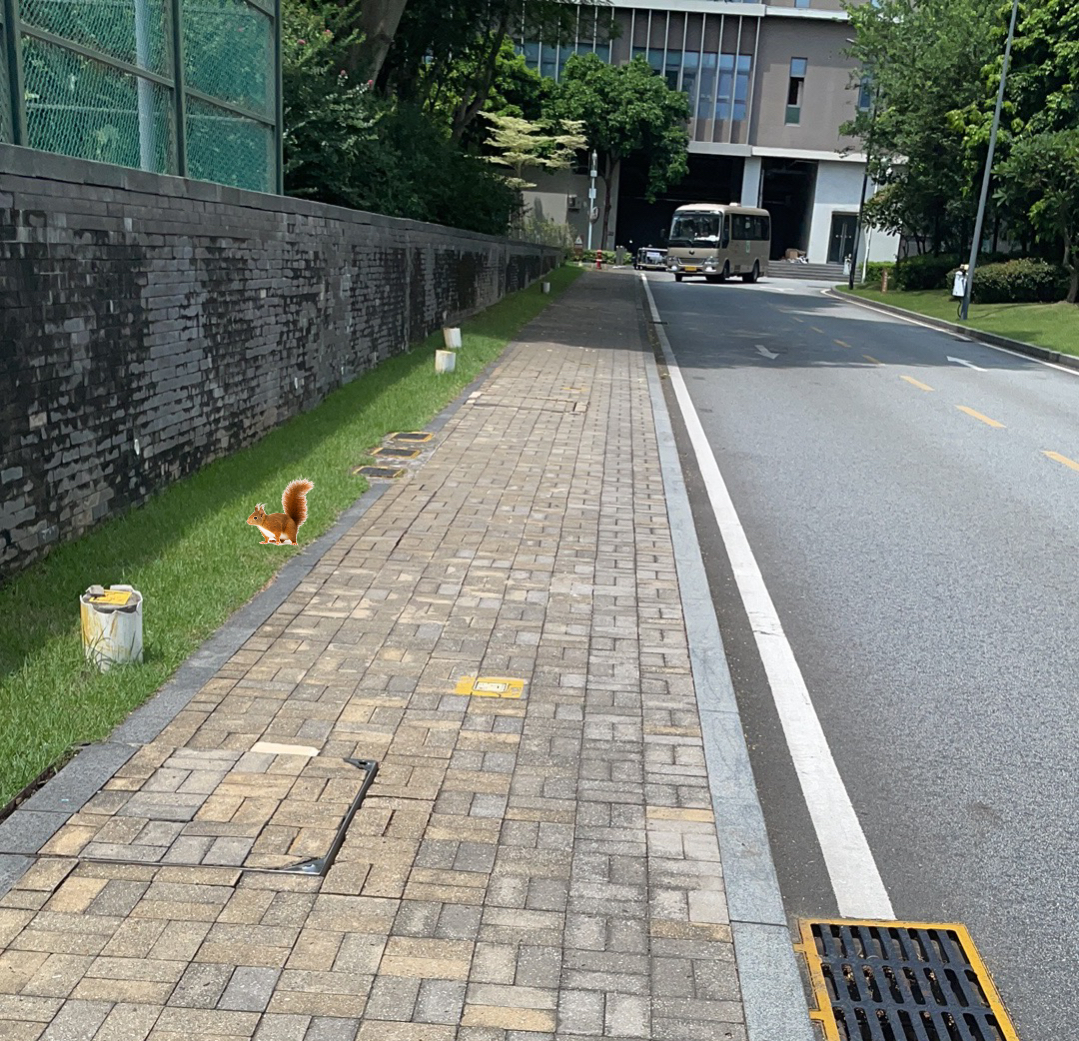} & Determine if this photo has been edited and provide a detailed explanation. & \textbf{1.} Confirm the photo is edited by identifying the added squirrel. \textbf{2.} Evaluate the explanation of how the squirrel's integration indicates editing. \textbf{3.} Assess the consistency of lighting and shadows related to the squirrel.\\
\hline
\begin{tabular}{@{}c@{}c@{}} Coding \\ Capability \\ with Vision\end{tabular} & \centering\arraybackslash\includegraphics[height=0.7in]{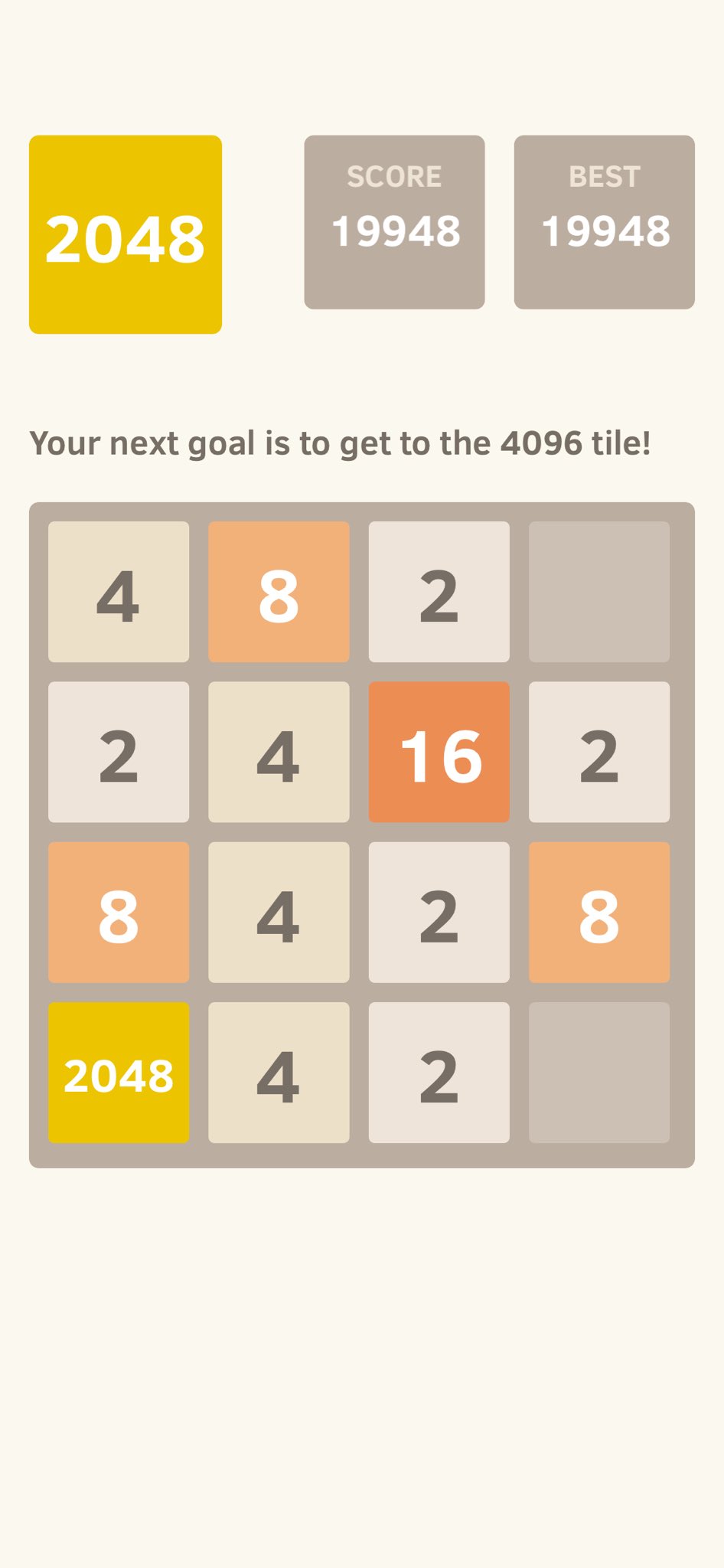} & The image depicts a game. You need to analyze the gameplay and use code to implement an identical game. & \textbf{1.} Identify game as sliding puzzle with numbered tiles; mention grid and tile combination mechanics. \textbf{2.} Describe the gameplay objective to combine tiles for a higher-numbered tile. \textbf{3.} Include code implementation reflecting game rules and sliding tile mechanism. \\ \hline
\end{tabular}}
\end{table*}

\subsection{Data Quality Control}

\begin{wrapfigure}{r}{0.45\textwidth}
\vspace{-40pt}

    \centering 
    \includegraphics[width=0.45\textwidth]{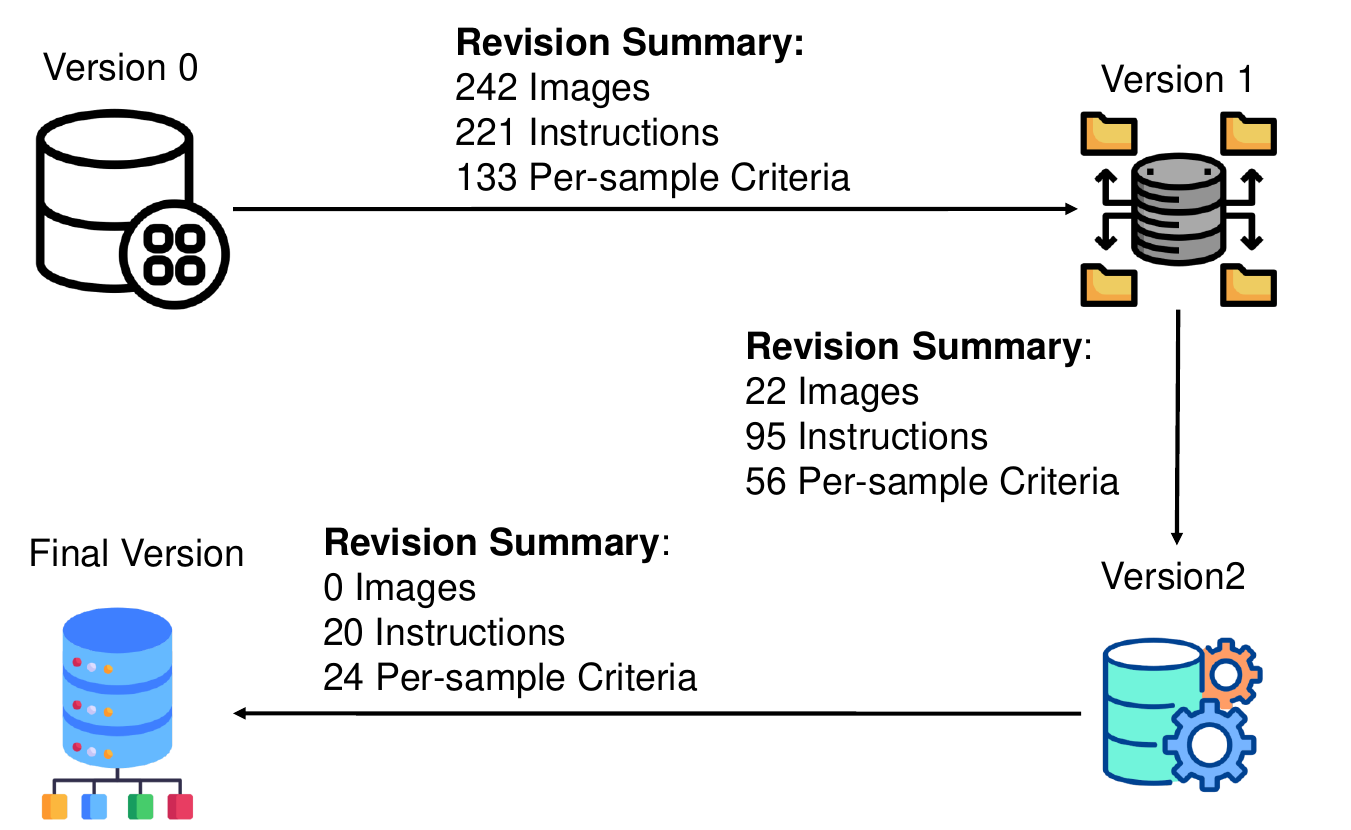} 
\caption{\label{fig:revision}Number of revised samples during the cross-review process.}
\vspace{-20pt}
\end{wrapfigure}

We employ a two-step protocol to validate the data: \textit{cross-review} and \textit{expert verification}.

\noindent\textbf{Cross-review~} Upon completion of data collection, the six volunteers are required to reviewe each other's work following the criteria in  Table~\ref{tab:guideline} in Appendix~\ref{sec:Guideline}.
Samples that did not meet the criteria were discarded and replaced to meet the required quota for each category. We conducted three rounds of cross-review, as shown in Figure~\ref{fig:revision}.



\noindent\textbf{Expert Verification~} Following cross-verification, an experienced volunteer with expertise in data evaluation inspects each sample for quality assurance. Low-quality samples are recollected using a process similar to cross-verification.






%

\section{Benchmarking}


\subsection{Evaluation Protocols}

\label{sec:evaluation_setting}

\textbf{Evaluation Settings~}Our evaluations are conducted in a \textit{ pairwise voting} manner. As illustrated in Figure~\ref{tab:exmp_pair_vote}, we adopt the superb GPT-4V\footnote{We use \texttt{gpt-4-1106-vision-preview} in December, 2023.} as the judge to vote for which answer is better given an image, a question and a pair of answers.
Each pair of answers consists of a benchmarked answer from an MLLM to be judged, and an anchor answer from LLaVA-v1.5-13B, which is a moderately strong MLLM. 
We then compute the number of \textit{win/tie/lose} of a benchmarked model over the anchor model.
Previous works~\citep{wang2023large,wu2023style,chen2024humans} investigate the positional bias in text-only evaluation. In this work, we also evidence that such bias also exists in multi-modal evaluation using GPT-4V (see App.~\ref{sec:postional_bias}), albeit subtle. 
To mitigate positional bias, we shuffle the position of each paired sample for all settings.




\textbf{Evaluation Prompts~} The evaluation criteria are based on the responses' quality, relevance to the image, as well as the given per-sample criteria. 
The evaluator is to choose which of the two answers is superior, except in two specific situations where a decision cannot be made: \textbf{1)} if the question is close-ended and both answers are equally correct or incorrect, or \textbf{2)} if both answers have significant factual inaccuracies or ethical concerns. 
In such cases, the evaluator can indicate their inability to decide, specifying the situation. 
In the last line, the judge should output a concise declaration of which answer is better or an indication that a decision cannot be made, citing the appropriate reason. Figure~\ref{fig:binary} in Appendix~\ref{sec:4V_detail} shows the evaluation prompt.
\textbf{Benchmarked MLLMs~}
We select 21 MLLMs to be evaluated based on two factors: popularity and architectural diversity. 
Primarily, the models chosen are some of the most widely used MLLMs at the time of this study, substantiated by their number of GitHub stars. 
Secondly, these models offer a broad spectrum of MLLM architectures, allowing us to conduct a comprehensive and comparative study.
Among these models, \textbf{GPT-4V}, \textbf{Claude-3} and \textbf{Gemini-Pro} are the cutting-edge proprietary models.
For open-sourced MLLMs, \textbf{LLaVA-series} and their variants (\textbf{LVIS}, \textbf{MiniGPT-v2}, \textbf{Yi-VL}, \textbf{kosmos2} and \textbf{ALLaVA}) offer a smart solution for modality adaptation. 
\textbf{BLIP-2}, \textbf{Qwen-VL}, \textbf{InstructBLIP}, \textbf{Cheetor} and \textbf{mPLUG-Owl2 }use query tokens as the bridge of visual and texual representation. 
\textbf{OpenFlamingo} uses a perceiver resampler to encode images and is the pioneer for multimodal In-Context Learning.
\textbf{LWM} is a recent MLLM that supports 1 million context.
\textbf{SEED-LLaMA} adopts VQ-based encoder, while \textbf{Fuyu} only uses a single linear layer to link two modalities.
See details  in App.~\ref{sec:models}.





\begin{table*}[t]
\footnotesize
\centering
\caption{Number of wins/ties/loses for each model over \textbf{LLaVA-v1.5-13B} (\textit{anchor}) on each level. Models are sorted by overall win rates in descending order. 
}
\vspace{1em}
\resizebox{\textwidth}{!}{
\setlength{\tabcolsep}{3pt} 
\begin{tabular}{@{}lrrrrrrc@{}}
\toprule
\textbf{Models} & \textbf{Perception} & \textbf{Understanding} & \textbf{Applying} & \textbf{Analyzing} & \textbf{Evaluation} & \textbf{Creation} & \textbf{Win Rates } \\ 
\midrule
Claude-3 & 56/13/1 & 98/9/3 & 45/11/4 & 83/14/3 & 33/5/2 & 33/6/1 & 0.83  \\
GPT-4V & 56/10/4 & 92/9/9 & 40/17/3 & 84/11/5 & 32/2/6 & 33/5/2 & 0.80  \\
LLaVA-v1.6-34B & 46/17/7 & 78/22/10 & 36/15/9 & 61/28/11 & 33/3/4 & 24/10/6 & 0.66 \\
LLaVA-v1.6-Vicuna-13B & 40/21/9 & 65/33/12 & 35/19/6 & 51/26/23 & 33/5/2 & 27/9/4 & 0.60 \\
\rowcolor{lightgray}
\multicolumn{7}{c}{LLaVA-v1.5-13B (\textit{anchor})} & 0.50 \\
LLaVA-v1.6-Vicuna-7B & 31/25/14 & 56/37/17 & 26/23/11 & 40/31/29 & 22/10/8 & 19/10/11 & 0.46 \\
ALLaVA-3B-Longer & 22/21/27 & 57/30/23 & 23/17/20 & 44/30/26 & 16/10/14 & 17/12/11 & 0.43 \\
Gemini-1.0-Pro & 45/10/15 & 36/35/39 & 24/19/17 & 33/28/39 & 9/8/23 & 16/8/16 & 0.39  \\
Qwen-VL-Chat & 34/22/14 & 38/36/36 & 26/18/16 & 35/29/36 & 15/6/19 & 9/12/19 & 0.37  \\
LVIS & 22/28/20 & 32/39/39 & 11/27/22 & 33/36/31 & 14/9/17 & 9/16/15 & 0.29  \\
mPLUG-Owl2 & 16/24/30 & 30/34/46 & 17/17/26 & 23/38/39 & 15/8/17 & 11/14/15 & 0.27  \\
LLaVA-v1.5-7B & 19/22/29 & 27/47/36 & 13/29/18 & 21/43/36 & 9/14/17 & 8/13/19 & 0.23 \\
MiniGPT-v2 & 12/25/33 & 24/32/54 & 11/25/24 & 17/38/45 & 9/9/22 & 6/6/28 & 0.19  \\
InstructBLIP & 15/16/39 & 13/36/61 & 6/23/31 & 13/29/58 & 10/7/23 & 4/9/27 & 0.15  \\
Cheetor & 12/20/38 & 7/27/76 & 10/22/28 & 16/23/61 & 4/4/32 & 3/4/33 & 0.12 \\
SEED-LLaMA & 16/15/39 & 5/25/80 & 10/21/29 & 7/25/68 & 3/7/30 & 3/3/34 & 0.10  \\
kosmos2 & 6/22/42 & 6/18/86 & 6/15/39 & 10/20/70 & 1/4/35 & 2/3/35 & 0.07  \\
Yi-VL-6B & 4/17/49 & 8/22/80 & 5/27/28 & 5/29/66 & 3/9/28 & 3/9/28 & 0.07 \\
Fuyu-8B & 7/19/44 & 7/27/76 & 6/14/40 & 4/22/74 & 3/7/30 & 0/6/34 & 0.06  \\
LWM & 2/18/50 & 5/15/90 & 4/21/35 & 2/18/80 & 3/2/35 & 2/6/32 & 0.04 \\
OpenFlamingo & 8/13/49 & 2/8/100 & 3/14/43 & 2/21/77 & 1/2/37 & 1/5/34 & 0.04 \\
BLIP2 & 3/13/54 & 2/15/93 & 6/8/46 & 0/22/78 & 0/1/39 & 0/2/38 & 0.03  \\

\bottomrule
\end{tabular}}

\label{tab:vote_llava}

\end{table*}

\vspace{-2mm}
\subsection{Evaluation Results}



Table~\ref{tab:vote_llava} presents the results of benchmarked models. The outcomes are quantified as the number of wins, ties, and losses for each model relative to LLaVA-v1.5, with the models sorted by their total win rates in descending order.  Claude-3 leads with a win rate of 0.83, significantly outperforming the second-place GPT-4V. LLaVA-v1.6-34B and LLaVA-v1.6-Vicuna-13B follow closely, both with win rates exceeding 0.6. LLaVA-v1.6-Vicuna-7B and ALLaVA-3B-Longer surpass Gemini-Pro in rankings, while Qwen-VL-Chat slightly underperforms compared to Gemini-Pro. These open-source models indicate that the gap with closed-source counterparts is narrowing, with smaller models like ALLaVA-3B-Longer also showing strong potential.

LVIS, mPLUG-Owl2, and LLaVA-v1.5-7B perform comparably to the anchor model, while MiniGPT-v2, InstructBLIP, Cheetor, and SEED-LLaMA show average performance.  kosmos-2, Yi-VL-6B, Fuyu-8B, LWM, OpenFlamingo, and BLIP2 all have win rates below 10\%, indicating poor performance, possibly due to a lack of instructional data during training or severe hallucinations.
\section{Validating Evaluation Paradigm and MLLM-Bench}
Affirming the validity of our proposed evaluation paradigm involves three  experiments which assess GPT-4V's alignment with human evaluations (in Section~\ref{sec:alignment_with_human}), the impact of per-sample criteria (in Section~\ref{sec:ablation_per_sample_criteria}), and the consistency of different MLLM judges (in Section~\ref{sec:ablation_mllm_judges}). Additionally, Section~\ref{sec:correlation_with_other_benchmarks} discusses the correlation between MLLM-Bench and other benchmarks.

\subsection{Alignment with Human Evaluation}
\label{sec:alignment_with_human}
\noindent\textbf{Settings}
For this alignment experiment, we recruit four undergraduate volunteers, all of whom are students at a university with an all-English curriculum. 
Each volunteer is paid according to the local salary (i.e., equivalent to roughly 10 dollars per hour). For volunteers participating in human evaluations, 
we sample 252 evaluation items. Each item in the evaluation set comprises a question associated with an image and two answers generated by models: one from the model under evaluation and the other from the anchor model LLaVA-v1.5-13B. We present the details of sample and model selection in App.~\ref{sec:pair_selection}.

To check the consistency between human evaluators and GPT-4V, we calculate the agreement between each individual's results as well as the aggregated results of human evaluators and GPT-4V. The aggregated results are derived from a majority vote among four human evaluators. This step combines individual judgments into a consensus and addresses situations like indecision or equal preference. If selections for the evaluated model and the anchor model are equal, or if more evaluators choose "\textit{unable to decide}" than any specific model, the outcome is classified as "\textit{unable to decide}".

\noindent\textbf{Results}
The last row of Table~\ref{tab:baseline_and_human_alignment} shows a high alignment between human evaluators and GPT-4V, with the agreement between individual evaluators and GPT-4V reaching 86.59\% and the agreement between the aggregated results of human evaluators and GPT-4V reaching 88.02\%. This high degree of alignment reaches the same level of agreement among human evaluators. The results validate the premise that GPT-4V can effectively mirror human judgment in the setting of MLLM evaluation, confirming GPT-4V as a reliable and effective tool in the evaluation process.


\begin{table}[t]
\footnotesize
\centering
\caption{Alignment between human evaluation and different evaluation methods.
\textit{det}: detection; \textit{cap}: caption. 
The \textbf{Aggregated} column aggregates the results of 4 human evaluators.}
\setlength{\tabcolsep}{3pt} 
\centering
\begin{tabular}{lccccc}
\toprule
Methods & \textbf{Judge 1} & \textbf{Judge 2} & \textbf{Judge 3} & \textbf{Judge 4}  & \textbf{Aggregated}\\ \midrule
GPT-4 + \textit{det} w/o  Criteria & 79.10\% & 78.20\% & 71.54\% & 72.48\% & 75.37\% \\
GPT-4 + \textit{det} w/  Criteria & 82.71\% & 78.79\% & 72.87\% & 80.18\% & 81.20\% \\ \midrule
GPT-4 + \textit{cap} w/o  Criteria & 80.00\% & 81.72\% & 74.59\% & 82.14\% & 81.15\% \\
GPT-4 + \textit{cap} w/  Criteria & 78.72\% & 80.32\% & 74.58\% & 80.98\% & 80.53\% \\ \midrule
GPT-4 + \textit{det}+ \textit{cap} w/o  Criteria & 81.12\% & 83.42\% & 75.27\% & 84.12\% & 83.00\% \\ 
GPT-4 + \textit{det}+ \textit{cap} w/  Criteria & 80.41\% & 82.29\% & 74.86\% & 83.54\% & 83.16\% \\ \midrule
GPT-4V w/o  Criteria & 81.62\% & 82.22\% & 75.42\% & 82.05\% & 82.80\% \\
\textbf{GPT-4V w/  Criteria (ours)} & \textbf{85.37\%} & \textbf{86.59\%} & \textbf{76.88\%} & \textbf{82.55\%} & \textbf{88.02\%} \\ \hline


\end{tabular}

\label{tab:baseline_and_human_alignment}

\end{table}

\subsection{Ablation on Per-sample Criteria}
\label{sec:ablation_per_sample_criteria}

\paragraph{Settings} 
To further demonstrate the superiority of our method, we include some baselines for comparison. 
  \textbf{GPT-4 + detection} \footnote{Details of bounding box generation are in App.~\ref{sec:bbox_generation}.} (w/ or w/o criteria): For each image, we adopt Detic~\citep{zhou2022detecting} generate top-10 BBoxes ranked by confidence score. Then we feed GPT-4 with BBox coordinates for evaluation.
 \textbf{GPT-4 + caption}\footnote{The prompt for caption generation with GPT-4V is in App.~\ref{sec:prompt_caption_generation}.} (w/ or w/o criteria):
    For each image, we generate a detailed caption using GPT-4V. Then we feed GPT-4 with the captions for evaluation. 
 \textbf{GPT-4 + detection + caption (w/ or w/o  criteria)}: For each image, top-10 BBox coordinates and a detailed caption are sent to GPT-4 in a single prompt for evaluation.
 \textbf{GPT-4V and \textbf{GPT-4V + criteria (ours)}}: Identical setting as in Section~\ref{sec:alignment_with_human}.
Prompts for GPT-4 evaluation are in App.~\ref{sec:prompts_gpt4_eval}.

\paragraph{Results}
Table~\ref{tab:baseline_and_human_alignment} summarizes the results. 
Per-sample criteria brings more than 5\%  overall agreement increase on GPT-4+\textit{det}  and GPT-4V evaluation, demonstrating the effectiveness of our proposed method.
Evaluation using GPT-4 with compound information (\textit{det} + \textit{cap}) outperforms using either of them, suggesting that these two methods introduce distinct information to GPT-4. 
We also relate our work to a recent work~\citep{fu2024blink}, which finds that GPT-4 performs well on some MLLM benchmarks with dense captions as input. Nevertheless, GPT-4V-as-a-judge still outperforms the its  GPT-4-as under our setting in terms of agreement with human, which demonstrate the necessity of adopting GPT-4V for evaluation in MLLM-Bench.

\subsection{Effects of Different MLLM Judges}
\label{sec:ablation_mllm_judges}
\paragraph{Settings}
While we use GPT-4V for our experiments, we aim to show that
the framework is designed to be adaptable and can seamlessly incorporate other potent MLLM models. To prove this, we conduct the same setting in Section~\ref{sec:evaluation_setting}, except that we replace GPT-4V with Claude-3-opus as the evaluator. We only evaluate on a subset of models due to limited budget.

\paragraph{Results}
We show the detailed results of \textit{Claude-3-Opus} in App.~\ref{app:claude_res}. The resulting ranking has a Spearman correlation of 0.95 with GPT-4V's voting results, which demonstrates that even if we change the model evaluator, the evaluation results highly align with the original ones. This flexibility allows for the substitution of GPT-4V with other powerful models, iterating along with the rapidly evolving field of MLLMs.

\subsection{Correlation with Other Benchmarks}\label{sec:correlation_with_other_benchmarks}
We discuss the correlation between MME-Perception, MME-Cognition and MM-Vet and MLLM-Bench. Specifically, we compute the pairwise Spearman correlation of rankings. 
Table~\ref{tab:bench_corr} shows that MLLM-Bench has the highest correlation with MM-Vet, which is a comprehensive benchmark testing 6 core abilities of MLLMs with short-answers. 
On the other hand, the Perception and Cognition splits of MME evaluate MLLMs at a fundamental and an advanced level, respectively. Therefore, the latter split has a higher correlation with MLLM-Bench than the former does. 
Thus, we highlight that our benchmark is proposed to complement with previous works, testing MLLMs at multiple levels with open-ended questions.

\begin{table}[ht]
\footnotesize
\centering
\caption{Pairwise Spearman correlation between different benchmarks: MME$^P$: MME-Perception; MME$^C$: MME-Cognition; MM-Vet; \textbf{MLLM-Bench}.}
\setlength{\tabcolsep}{3pt} 
\centering
\begin{tabular}{lcccc}
\hline
             & MME$^P$ & MME$^C$ & MM-Vet & \textbf{MLLM-Bench} \\ \hline
MME$^P$      & 1.00    & 0.14    & -0.40  & 0.57               \\
MME$^C$      & 0.14    & 1.00    & 0.80   & 0.81               \\
MM-Vet       & -0.40   & 0.80    & 1.00   & 0.89               \\
\textbf{MLLM-Bench}    & 0.57    & 0.81    & 0.89   & 1.00        \\
\hline
\end{tabular}

\label{tab:bench_corr}
\end{table}

\vspace{-2mm}

\section{Conclusion}
In this paper, we propose a new paradigm for MLLM evaluation and present MLLM-Bench, a benchmark for automatically evaluating the MLLMs' ability on open-ended queries.
It is derived from a comprehensive taxonomy, paving the way for a more responsible and conscientious approach to AI development. In MLLM-Bench, instead of providing the judge model with a standard answer, we equip each evaluation sample with evaluation criteria to evaluate open-ended answers reasonably. 
Experimental analysis on MLLM-Bench shows that evaluation using a potent MLLM using our per-sample criteria strategy aligns better with human than other baselines.
We hope that MLLM-Bench can introduce more insights to MLLM evaluation and development.

\newpage

\bibliographystyle{neurips_data_2024} 
\bibliography{sample-base}

\appendix

\clearpage
\section{Detailed taxonomy}
\FloatBarrier

\label{sec:taxonomy}
\begin{table}[ht]
    \centering
    \footnotesize
    \caption{Overview of 42 capabilities on 6 cognitive levels in MLLM-Bench.}
    \begin{tabular}{llc}
        \toprule
        Capability Level & Capability  \\
        \midrule
        \multirow{7}{*}{Level 1: Perception} & General Object Recognition \\
        & OCR \\
        & Multilingual Text Recognition \\
        & Action Recognition \\
        & Symbol Recognition \\
        & Food Recognition \\
        & Landmark Recognition \\
        
        \midrule
        \multirow{11}{*}{Level 2: Understanding} & Scene Understanding \\
        & Attribute Recognition \\
        & Image Topic Understanding \\
        & Hidden Objects Recognition \\
        & Facial Expression Recognition \\
        & Emotion Understanding \\
        & multimodal Commonsense Understanding \\
        & Joke and Meme Understanding \\
        & Multilingual Multicultural Understanding \\
        & Document Understanding \\
        & Table Understanding \\
        \midrule
        \multirow{6}{*}{Level 3: Applying} & Object Localization \\
        & Object Counting \\
        & Spatial Relationship Understanding \\
        & Medical Image Understanding \\
        & Image Captioning \\
        & Dense Captioning \\
        \midrule
        \multirow{12}{*}{Level 4: Analyzing} & Natural Relation Understanding \\
        & Chart Understanding  \\
        & Attribute Comparison \\
        & Difference Finding \\
        & Event Cause Reasoning \\
        & Social Relation Reasoning \\
        & Identity Reasoning \\
        & Function Reasoning \\
        & Physical Property Reasoning \\
        & Visual Math Reasoning \\
        & Action Prediction \\
        & Trend Prediction \\
        \midrule
        \multirow{4}{*}{Level 5: Evaluation} & Image Quality Evaluation \\
        & Damage Evaluation \\
        & Fake Image Detection \\
        & Ethical Problem Detection \\
        \midrule
        \multirow{2}{*}{Level 6: Creation} & Coding Capability with Vision \\
        & Visual Storytelling \\
        \bottomrule
    \end{tabular}

\label{tab:detailed_taxonomy}
\end{table}

\FloatBarrier

The detailed taxonomy is shown in Table~\ref{tab:detailed_taxonomy}.
The table provides a comprehensive overview of 42 capabilities distributed across six cognitive levels within the MLLM-Bench framework. These capabilities are designed to benchmark the performance of multimodal language and vision models (MLLMs) across a range of tasks that mimic human cognitive abilities. Here's a summary of the capabilities by level:

\begin{itemize}
    \item \textbf{Level 1: Perception} - This level focuses on basic recognition tasks such as identifying objects, symbols, actions, and landmarks, as well as recognizing text in various languages and formats.

    \item \textbf{Level 2: Understanding} - At this level, capabilities extend to more complex comprehension tasks like scene and attribute recognition, understanding emotions and facial expressions, recognizing hidden objects, and grasping multimodal commonsense, including jokes and memes, across different languages and cultures.

\item \textbf{Level 3: Applying} - This involves applying knowledge to practical tasks, including localizing objects, counting, understanding spatial relationships, interpreting medical images, and generating image captions and dense captions that describe images in detail.

\item \textbf{Level 4: Analyzing} - Here, the focus shifts to analytical tasks such as understanding natural relations, analyzing charts, comparing attributes, finding differences, reasoning about events and social relationships, deducing identity and function, reasoning about physical properties, and predicting actions and trends.

\item \textbf{Level 5: Evaluation} - This level assesses the model's judgment capabilities, including evaluating image quality, assessing damage, detecting fake images, and identifying ethical issues.

\item \textbf{Level 6: Creation} - The highest level of cognitive capability involves creative tasks, such as using vision to aid in coding and telling stories visually.

\end{itemize}

Each level builds on the previous ones, progressing from basic perception to complex and creative problem-solving, reflecting an ascending order of cognitive complexity and capability required by MLLMs.

\section{More Details on Data Collection and Annotation}
\label{sec:details_annotation}

We mainly introduce the division of labor for these six volunteers as well as the pipeline for data collection and annotation.

\subsection{Task Distribution for Volunteers}
\label{sec:task_distribution}

To ensure consistency and quality across our dataset, we have engaged six volunteers, each of whom is tasked with collecting and annotating data within one or two specific levels. The distribution of responsibilities is as follows:

\begin{itemize}
   \item  Perception Level: One volunteer is responsible for all 70 instances.
    \item  Understanding Level: Two volunteers share this category, with each annotating 55 instances.
    \item  Applying Level: A single volunteer manages all 60 instances.
    \item  Analyzing Level: This is shared between two volunteers, with each handling 60 instances.
    \item  Evaluation and Creation Levels: A single volunteer is responsible for the combined total of 60 instances across these two categories to balance the workload.
\end{itemize}

This division ensures that the workload is approximately equal for each volunteer. The entire annotation cycle was completed over a span of 21 days.

\subsection{Guideline}
\label{sec:Guideline}

The guideline for data collectors outlines essential considerations for data annotation to ensure the dataset's quality and relevance, see Table~\ref{tab:guideline}. Firstly, data currency is emphasized by prioritizing the most recent images, ideally within three months, sourced from social networks or daily life captures to mitigate the risk of data leakage. License-friendliness is crucial, advocating for the use of publicly licensed data with clear sharing agreements to respect intellectual property rights. Image clarity is necessary, with a recommendation for a mix of high and lower resolution images (with a minimum short-edge resolution of 512) to accurately represent real-world conditions. Impartiality is maintained by avoiding content related to sensitive topics, ensuring the dataset's neutrality and broad applicability. Instruction-image cohesion is vital, requiring precise and contextually reflective instructions to facilitate clear understanding and effective model testing. Lastly, diverse response formats are encouraged to enrich the dataset, promoting varied feedback beyond simple binary choices to better simulate real-world interactions and enhance analytical capabilities.

\begin{table*}[t]
\footnotesize
    \centering
    \caption{The guideline for data collectors that states key considerations for data annotation.}
    \vspace{1em}
    \begin{tabular}{m{2.35cm}p{11.5cm}}
\toprule
\textbf{Characteristics} & \multicolumn{1}{c}{\textbf{Description}} \\
\midrule
Data Currency & To mitigate the risk of data leakage, it is crucial to utilize the most current images available. Annotators should \textbf{prioritize images sourced from social networks like Twitter or directly captured in daily life}, ideally within three months prior to our data collection phase. This approach helps prevent the collected data from being previously used in the training of evaluated models. Maintaining and updating the dataset in real time is recommended to address potential data leakage concerns effectively. \\
\midrule
License-friendliness & When selecting data sources, prioritize those that are \textbf{publicly licensed and offer favorable sharing agreements}. Social networks like Twitter or content personally photographed by annotators are preferred due to their clear copyright status. This ensures that our data collection respects intellectual property rights and adheres to legal requirements. \\
        \midrule
Image Clarity & The collected images must be of sufficient quality to be identifiable by humans. However, incorporating a selection of lower-resolution images \textbf{(the short-edge resolution is at least 512)} is also advised to accurately represent the diversity of data quality encountered in real-world scenarios. This approach ensures that our dataset reflects practical conditions and challenges.  \\
        \midrule
Impartiality & To maintain the dataset's neutrality, \textbf{avoid including content related to sensitive topics} such as geopolitical issues. This commitment to impartiality ensures that our dataset can be used widely without bias, supporting a broad range of applications and studies. \\
\midrule
Instruction-Image Cohesion & To ensure images and instructions are related, the instructions should not only be precise, but also be tailored to reflect the context depicted in the image as well as to test the specific model capability. 
For this to be effective, the language used in the instructions should be precise and unambiguous, facilitating a clear understanding of the image content and task requirements.\\ \midrule
Diverse Responses \;\; Formats & Encouraging a variety of response formats enriches the dataset and better simulates real-world interactions. Instead of limiting responses to simple binary choices, data collectors are encouraged to seek out descriptive narratives and diverse forms of feedback. This strategy distinguishes our dataset from existing benchmarks and enhances its applicability to complex analytical tasks. \\
\bottomrule
    \end{tabular}

    \label{tab:guideline}
\end{table*}

\section{Data Statistics}
\label{sec:d_sta}

\begin{figure}[ht]
    \centering 
    \includegraphics[width=0.5\textwidth]{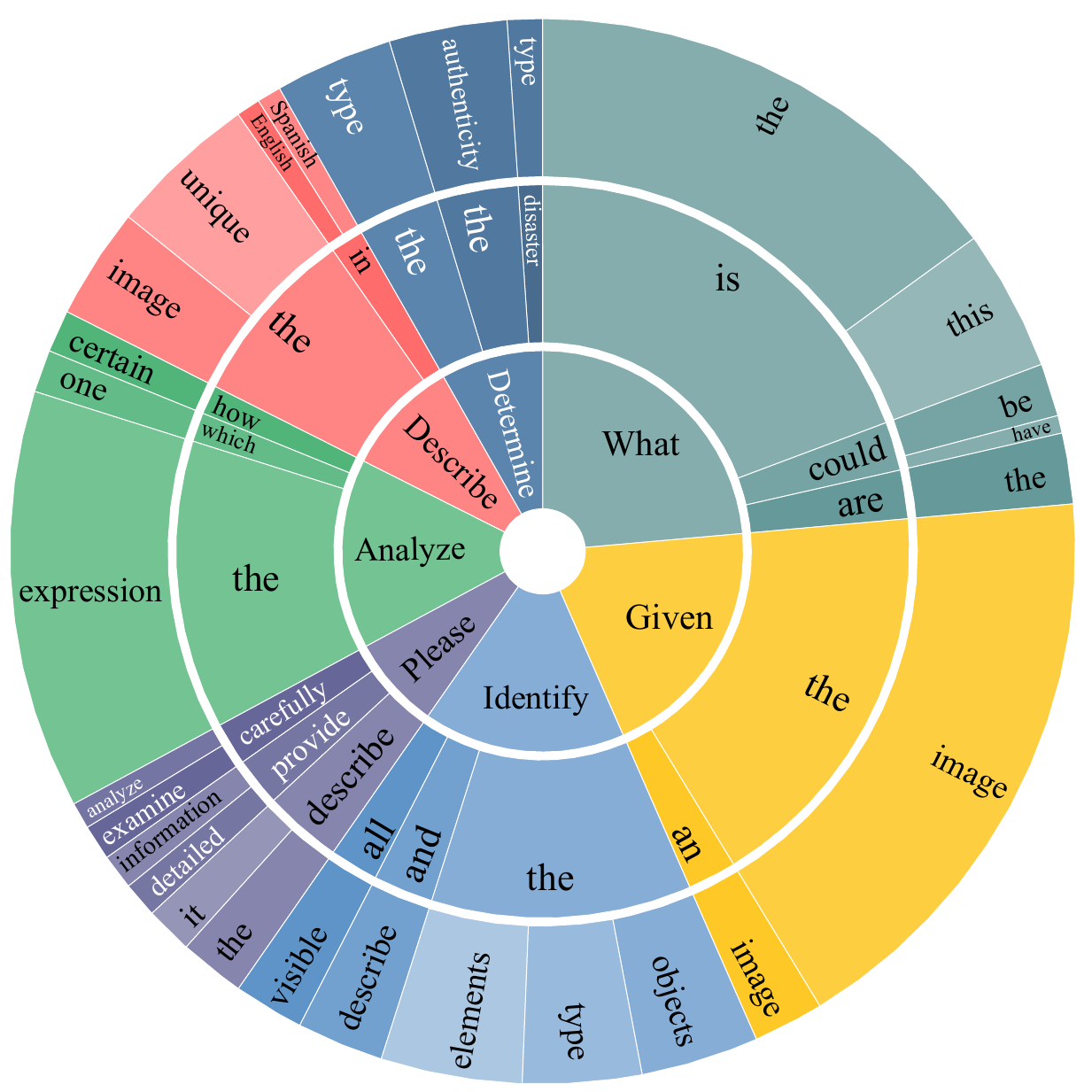} 
    \caption{\label{wordcloud}Distribution of questions in MLLM-Bench. We present the relative distribution of these recurring Instructions and their subsequent distributions.}
\end{figure}

MLLM-Bench is characterized by a rich diversity and complexity of instructions, each tailored to probe a specific capability of multimodal large language models. These instructions challenge models to generate responses that are both comprehensive and descriptive, engaging with the multifaceted nature of real-world scenarios and information. To illustrate the breadth of our instruction set, we present a word cloud visualization that encapsulates the frequency of terms within our instructions, as shown in Figure~\ref{wordcloud}.
We list one example per category in Table~\ref{tab:sample}. One can view and download our data and codes from \url{https://anonymous.4open.science/r/MB-5F39}.

\section{Details of Benchmarked Models}
\label{sec:models}


Table~\ref{tab:model_architecture} presents a comparison of various models, highlighting their characteristics such as their open-source availability and architectural components, including visual adapters and base large language models (LLMs). Models vary in size from 1.7B to 14B parameters, with some details remaining confidential, indicated by "/". Not all models are open-sourced, as exemplified by GPT-4V. Architectural details vary, with some models utilizing visual adapters like CLIP-ViT-L, ViT+Q-Former, and others, while the base LLMs mentioned include Vicuna-13B, LLaMA2-7B, and more. The table aims for clarity by abbreviating model names and provides a snapshot of the diverse approaches in integrating visual processing with language models, demonstrating a range of strategies for enhancing model capabilities.


\FloatBarrier
\begin{table*}[htbp]

\centering
\caption{Model architecture and popularity. "/" means the model either uses a private download link or their download counts on HuggingFace are not shown.}
\vspace{1em}
\resizebox{1\textwidth}{!}{
\begin{tabular}{lccc}
\toprule

\textbf{Models} &\textbf{ MLLM Architecture} & \textbf{GitHub Stars} & \textbf{Huggingface Download} \\
\midrule
\multicolumn{4}{c}{\textit{Closed-Source}} \\
\midrule
 GPT-4V~\citep{GPT4V} & \multicolumn{3}{c}{/}\\
Claude-3 ~\citep{Claude3} & \multicolumn{3}{c}{/} \\
Gemini-Pro ~\citep{Gemini} & \multicolumn{3}{c}{/}\\
\midrule
\multicolumn{4}{c}{\textit{Open-Source}} \\
\midrule
LLaVA-v1.5-13B~\citep{liu2023improved} & Pretrained Vision Encoder + Projector + LLM & 15.4K & 333.7K \\
LVIS-Instruct4v-LLaVA-7B~\citep{wang2023believe} & Pretrained Vision Encoder + Projector + LLM & 122 & 5 \\
MiniGPT-v2~\citep{chen2023minigptv2} & Pretrained Vision Encoder + Projector + LLM & 24.7K & / \\
LLaVA-v1.5-7B~\citep{liu2023improved} & Pretrained Vision Encoder + Projector + LLM & 15.4K & 703K \\
LLaVA-v1.6-Vicuna-7B~\citep{LLaVA-NeXT} & Pretrained Vision Encoder + Projector + LLM & 15.4K & 1.2M \\
LLaVA-v1.6-Vicuna-13B~\citep{LLaVA-NeXT} & Pretrained Vision Encoder + Projector + LLM & 15.4K & 100.1K \\
LLaVA-v1.6-34B~\citep{LLaVA-NeXT} & Pretrained Vision Encoder + Projector + LLM & 15.4K & 592.8K \\
Yi-VL-6B~\citep{Yi} & Pretrained Vision Encoder + Projector + LLM & 7K & 17.2K \\
ALLaVA-3B-Longer~\citep{ALLaVA} & Pretrained Vision Encoder + Projector + LLM & 134 & 93 \\
kosmos2~\citep{peng2023kosmos2} & Pretrained Vision Encoder + Grounded LLM & 18.1K & 29.2K \\
LWM~\citep{LWM} & Pretrained Vision Encoder + Projector + Long-Context LLM & 6.6K & / \\
BLIP2-Flan-T5-XL~\citep{li2023blip2} & Query tokens + LM & 8.5K & 35.4K \\
Qwen-Vl-Chat~\citep{bai2023qwenvl} & Query tokens + LLM & 3.4K & 289.9K \\
InstructBLIP-Vicuna-13B~\citep{dai2023instructblip} & Query tokens + LLM & 8.5K & 5.4K \\
mPLUG-Owl2~\citep{ye2023mplugowl2} & Query tokens + LLM with Modality-Adaptive Module & 1.9K & 9.7K \\
Cheetor~\citep{DEMON} & Query tokens + VPG-C + LLM & 308 & / \\
Fuyu-8B~\citep{fuyu-8b} & Linear Vision Encoder + LLM & / & 17.9K \\
SEED-LLaMA~\citep{ge2023making} & VQ-based Vision Encoder + LLM & 445 & / \\
OpenFlamingo~\citep{OpenFlamingo} & Perceiver Resampler + LLM with Gated Cross-Attention Layers & 3.4K & / \\

\bottomrule
\end{tabular}}


\label{tab:model_architecture}
\end{table*}

\begin{figure*}[htbp]
    \centering
    \includegraphics[width=\textwidth]{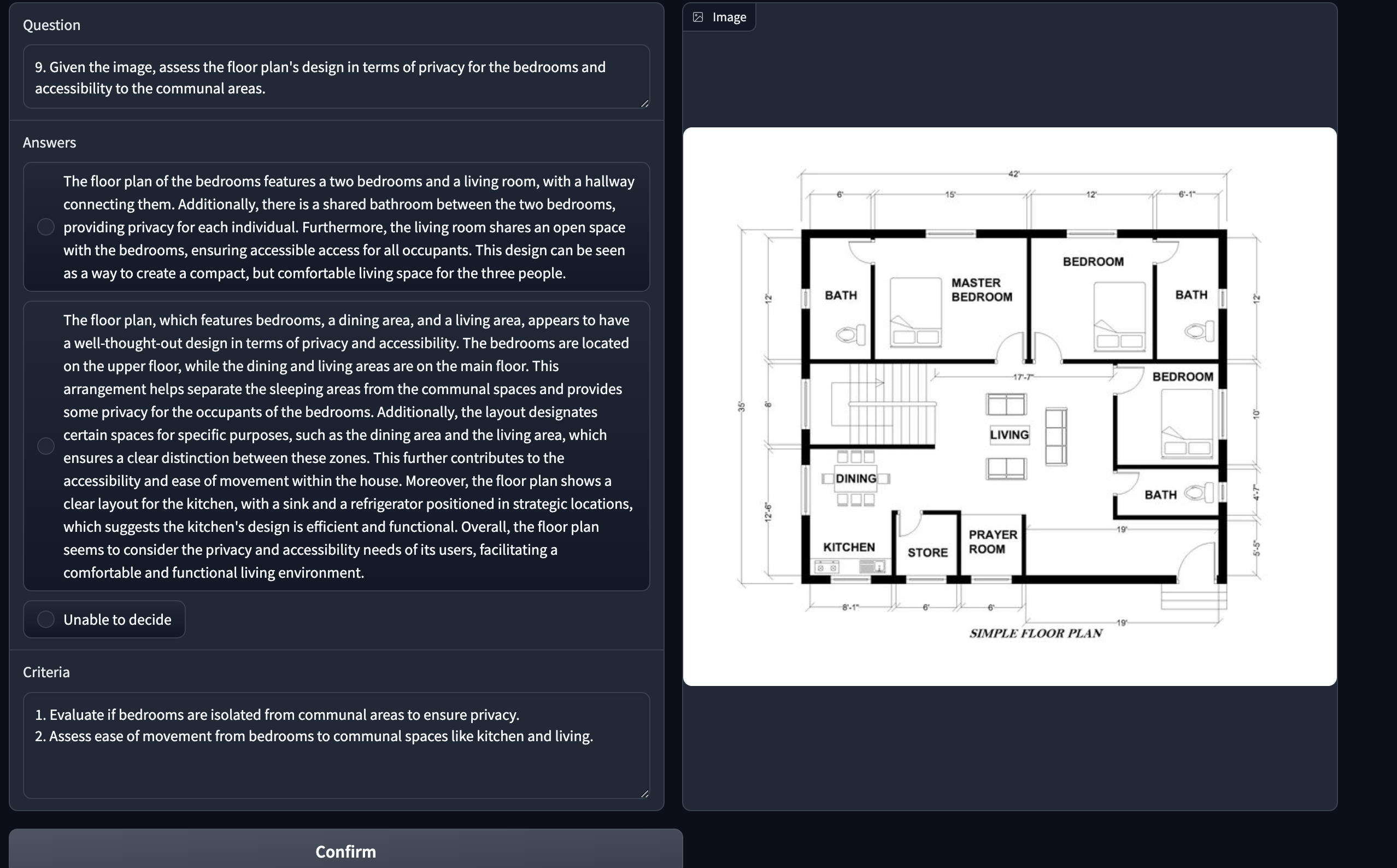}
    \caption{User interface of human evaluation.}
    \label{fig:human_eval_demo}
\end{figure*}




\section{Details of GPT-4V Evaluation}
\label{sec:4V_detail}
\subsection{Potential Deficiency of GPT-4V}
\label{sec:4v_eval}
While GPT-4V is a potential evaluator, its assessment outcomes may not always align perfectly with factual accuracy or human standards. There are situations where GPT-4V itself cannot handle the task. If GPT-4V itself cannot solve the problem, it cannot be a qualified judge to conduct the evaluation. Table~\ref{tab:4v_short} shows examples of GPT-4V's incorrect evaluation without our per-sample criteria.

\begin{table*}[htbp]
\footnotesize
\centering

{\renewcommand{\arraystretch}{0.6}
\caption{\label{tab:4v_short} \small Examples showing GPT-4V's capability fall short.} 
\vspace{1em}
\resizebox{\textwidth}{!}{
\begin{tabular}{p{0.9\columnwidth}p{0.3\columnwidth}}
\toprule
\multicolumn{2}{c}{\textbf{Question}: Determine the percentage increase in 'Total Income' from the quarter 30 Sep 2023 to the quarter 30 June 2023.} \\
\multicolumn{2}{c}{
\centering\arraybackslash\includegraphics[height=3in]{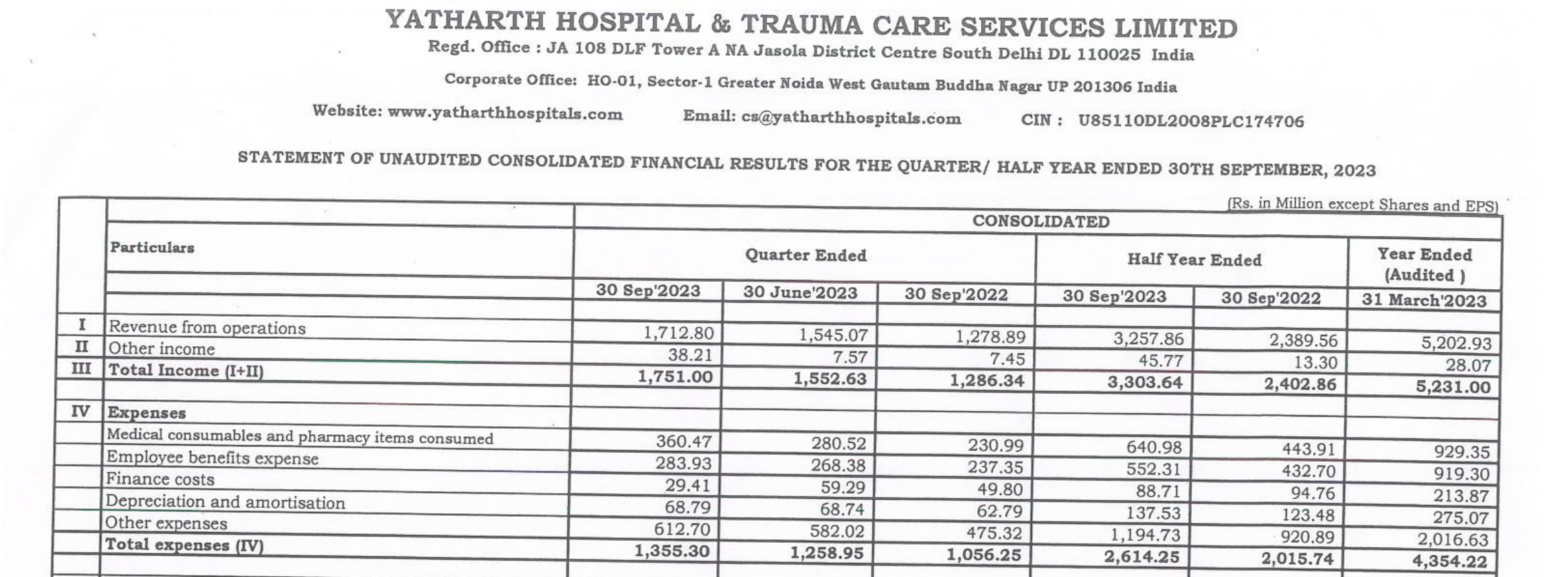} } \\
\midrule
\textbf{{\color{blue}Answer 1 ~(LLaVA-v1.5-13B):}} &  \textbf{{\color{red} Answer 2~(Fuyu-8B):}}  \\

{ 
To calculate the percentage increase in 'Total Income' from the quarter ended 30 Sep 2023 to the quarter ended 30 June 2023, we need to use the formula: (Current value - Previous value) / Previous value x 100. The Total Income values for the two quarters are as follows: 30 Sep 2023: 1,659,059 30 June 2023: 1,174,573 Using the formula, the percentage increase in Total Income is calculated as follows: (1,659,059 - 1,174,573) / 1,174,573 x 100 = 43.816\% So, there is a 43.816\% increase in Total Income from the quarter ended 30 Sep 2023 to the quarter ended 30 June 2023.} & {240.28} \\
\midrule
\multicolumn{2}{l}{\textbf{Criteria}:\textit{1. Correctly calculates percentage increase from 'Total Income' for specified quarters.}} \\
\multicolumn{2}{l}{\textit{2. Matches the percentage increase with the reference:\textbf{ 12.78\%}.}} \\
\multicolumn{2}{l}{\textit{3. Explains the calculation process or provides the correct formula used.}} \\
\midrule
\textbf{GPT-4V Judgment (\textit{w/o criteria})} & \textbf{GPT-4V Judgment (\textit{w/ criteria})} \\
\rowcolor{gray!20}\color{blue}\textbf{Answer 1} & \color{fg}\textbf{Unable to decide: situation one} \\

\bottomrule
\end{tabular}}}

\end{table*}

\subsection{Prompt for Pairwise Voting Using GPT-4V with Per-Sample Criteria}
Our evaluations are conducted following a \textit{ pairwise voting} protocol.  A stronger model is expected to have a larger number of wins in pairwise voting.
\begin{figure*}[htbp]
\centering
\footnotesize
\begin{AIbox}{Prompt for Pairwise Voting using GPT-4V}
{\bf Prompt:} \\
{\footnotesize
\#\#\# You are an excellent evaluator.\\
\#\#\# Your assignment involves providing evaluations for given responses.\\
\#\#\# Each evaluation consists of *an image*, *a question*, a *question type*, and *two corresponding answers*. Your task is to discern which answer is superior based on the **quality** and its alignment w.r.t the image.\\
\#\#\# There are only two situations where you may choose 'unable to decide':\\
\#\#\#\# Situation one: The question type is 'close-ended' and both answers are correct or wrong. \\
\#\#\#\# Situation two: Both answers contain considerable factual errors or ethical issues.\\
\#\#\# Otherwise, you should always choose a better answer by responding 'Answer1' or 'Answer2'.\\
\#\#\# You should ONLY output your vote 'Answer1', 'Answer2', 'unable to decide: situation one', or 'unable to decide: situation two' in the last line. \\
\textasciitilde\textasciitilde\textasciitilde Question \\
\{question\} \\
\textasciitilde\textasciitilde\textasciitilde \\
\textasciitilde\textasciitilde\textasciitilde Question Type \\
\{question\_type\} \\
\textasciitilde\textasciitilde\textasciitilde \\
\textasciitilde\textasciitilde\textasciitilde Answer1 \\
\{answer1\} \\
\textasciitilde\textasciitilde\textasciitilde \\
\textasciitilde\textasciitilde\textasciitilde Answer2 \\
\{answer2\} \\
\textasciitilde\textasciitilde\textasciitilde \\
\#\#\# Please refer to the given criteria when you making the judgment

Criteria: \{criteria\}



}
\end{AIbox} 
\caption{\label{fig:binary} The prompt used for Directing Voting using GPT-4V.  }
\end{figure*}
The prompt of conducting voting by GPT-4V is shown in Figure~\ref{fig:binary}.

\subsection{Anchors in Pairwise Evaluation}

To facilitate a fair and consistent comparison across multiple models, we employ an anchor-based evaluation strategy. For each protocol, we use answers from the moderately powerful model \textit{LLaVA-v1.5-Vicuna-13B} as benchmarks. 
\section{Additional Results for Validation Experiments}

\subsection{Correlation with Human Alignment}
\label{sec:pair_selection}

\paragraph{Model Selection} Our test set for the experiment is constructed by extracting two data points from each of the 42 capabilities identified in our benchmark. We select three models representing different levels of capabilities—top, middle, and bottom—as determined by GPT-4V's direct voting outcomes. These models are \textbf{Qwen-VL-Chat}, \textbf{InstructBLIP-Vicuna 13B}, and \textbf{BLIP2-Flan-T5-XL}. \textbf{ LLaVA-v1.5-13B} serves as an anchor. The evaluation set thus comprised 252 items in total. All volunteers are required to make judgments on all 252 data samples.

\paragraph{Interface for Human Evaluation} The user interface of human evaluation is shown in Figure~\ref{fig:human_eval_demo}.

\paragraph{Detailed Results}Table~\ref{tab:human_correlation} presents the pairwise agreement among four human evaluators, offering insights into inter-evaluator alignment levels regarding a set of evaluations. The inter-evaluator agreement is calculated by the matching percentage of two evaluators' voting results. The agreement ranges from 0.80 to 0.88, indicating a high degree of consensus among the evaluators. Specifically, the correlation values suggest that while there is a strong overall alignment in their assessments, each evaluator also brings a unique perspective to the evaluation process. Evaluator 1 and Evaluator 2 exhibit the highest correlation (0.88), suggesting their evaluations are most closely aligned. In contrast, the lowest correlation is observed between Evaluator 1 and Evaluator 3 (0.69), indicating a lesser, yet still significant, level of agreement. These findings underscore the evaluators' ability to consistently recognize and rank the evaluated items according to similar criteria, while also maintaining individual discretion in their judgments. This balance between consensus and individuality is crucial for ensuring both the reliability and the richness of the evaluation process, highlighting the evaluators' competence in providing nuanced assessments.

\subsection{Positional Bias of GPT-4V-as-a-Judge}
\label{sec:postional_bias}
\begin{figure}[htbp]
    \centering
    \includegraphics[width=\columnwidth]{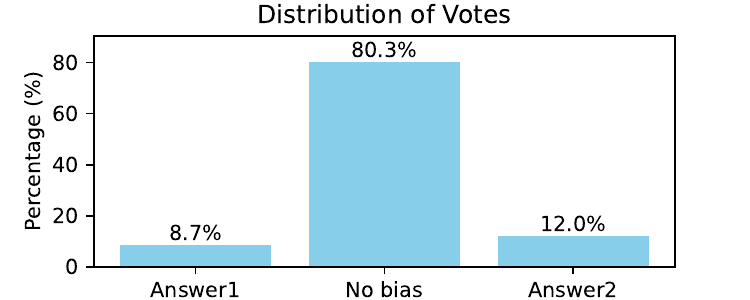}
    \caption{Distribution of votes from GPT-4V.}
    \label{fig:positional_bias}
\end{figure}
We investigate the positional bias of GPT-4V-as-a-judge. Specifically, if GPT-4V has consistent votes when orders of a pair of answers are shuffled, then we mark the vote as unbiased (No bias). Otherwise, we mark it as positionally biased towards the first (Answer 1) or the second answer (Answer 2). 
We summarize the results in Figure~\ref{fig:positional_bias}. 
Among the biased votes, GPT-4V shows a slight preference towards Answer 2.
However, the majority of the votes (80.3\%) are free from positional bias, manifesting the validity of using GPT-4V-as-a-judge. 

\subsection{Length Bias of GPT-4V-as-a-Judge}
\label{sec:length_bias}
\begin{figure}[htbp]
    \centering
    \includegraphics[width=\columnwidth]{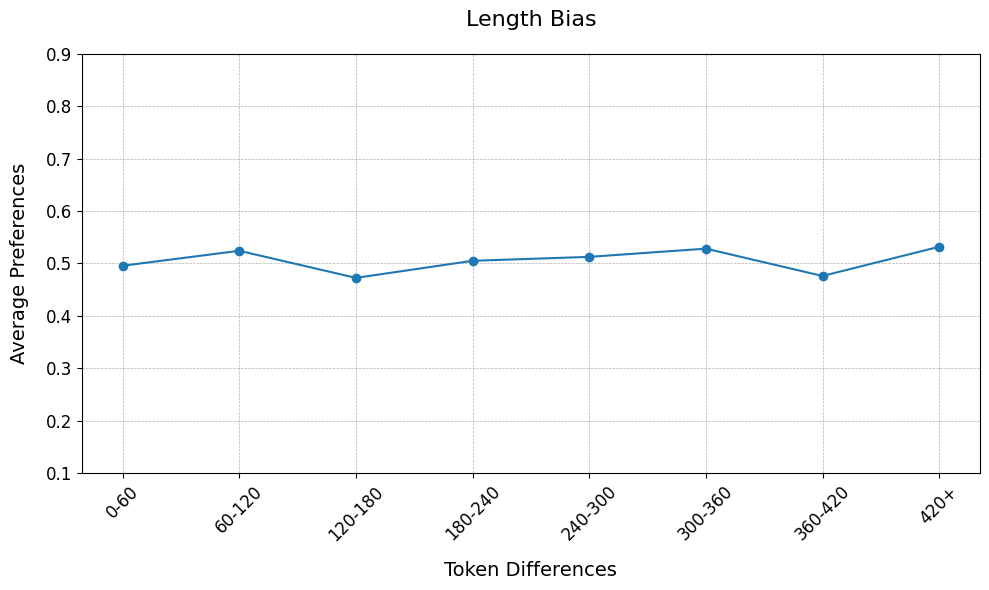}
    \caption{Preference towards different output lengths of GPT-4V.}
    \label{fig:length_bias}
\end{figure}
To investigate the potential correlation between the judge's decision and the generated length or text style of the model, we conducted a length bias experiment to investigate GPT-4V's preference towards different lengths of the output, as shown in Figure~\ref{fig:length_bias}. The y-axis of the figure shows the average preference of GPT-4V, where 0 represents a preference for short answers and 1 represents a preference for long answers. The x-axis shows the token differences between the two evaluated answers. The results indicate that GPT-4V's preference towards the generated length of the output is negligible.

\subsection{Reproducibility of Results} 
\label{sec:rep_res}
We select three models and test them three times using 'gpt-4-1106-vision-preview' (all other experiment settings stay the same as the voting experiment in Section~\ref{sec:evaluation_setting}), checking the consistency between results. Results are shown in Table~\ref{table:performance_metrics}. Each entry shows the average win rate and standard deviation. The results indicate a high consistency between different evaluations on the same model, demonstrating the robustness and reliability of using GPT-4V-as-a-judge.

\begin{table}[ht]
\centering
\caption{Performance metrics of different models across various capabilities.}
\resizebox{\textwidth}{!}{
\begin{tabular}{lccccccc}
\hline
\textbf{Models} & \textbf{Perception} & \textbf{Understanding} & \textbf{Applying} & \textbf{Analyzing} & \textbf{Evaluation} & \textbf{Creation} & \textbf{Total} \\ 
\hline
LLaVA-v1.6-34B & 0.62 (0.05) & 0.68 (0.03) & 0.59 (0.01) & 0.61 (0.00) & 0.82 (0.02) & 0.65 (0.04) & 0.65 (0.01) \\ 
Openflamingo & 0.12 (0.01) & 0.02 (0.01) & 0.05 (0.01) & 0.03 (0.01) & 0.03 (0.02) & 0.02 (0.01) & 0.04 (0.01) \\ 
Yi-VL-6B & 0.05 (0.01) & 0.06 (0.01) & 0.11 (0.02) & 0.06 (0.01) & 0.07 (0.01) & 0.08 (0.01) & 0.07 (0.00) \\ 
\hline
\end{tabular}
}
\label{table:performance_metrics}
\end{table}

\subsection{Bounding Box Generation}
\label{sec:bbox_generation}
We employ Detic\footnote{We use the version Detic\_LCOCOI21k\_CLIP\_SwinB\_896b32\_4x\_ft4x\_max-size.}~\citep{zhou2022detecting} to generate bounding box coordinates and their labels. For each image, we only keep the top-10 objects according to confidence scores. For each object, we store the coordinates of top-left and bottom-right corner, which will later be sent to text-only GPT-4.

\subsection{Prompts for Caption Generation}
\label{sec:prompt_caption_generation}
The prompt for generating image captions is shown in Figure~\ref{fig:prompt_caption}.

\begin{figure*}[htbp]
\footnotesize
\begin{AIbox}{Prompt for generating image captions using GPT-4V}

{\bf Prompt:} \\
{\footnotesize
\#\#\# You are an excellent image describer and captioner \\
\#\#\# Your task is to describe the given image as detailed as possible and give a caption for the image \\
\#\#\# Do not mention anything from the prompt in your response \\
\#\#\# You will follow the instructions to the best of your ability \\
\#\#\# Your response should follow the following format \\
<start of description> \\
\{description\} \\
<end of description>
}
\end{AIbox}
\caption{ \label{fig:prompt_caption}Prompts for generating image captions using GPT-4V.}
\end{figure*}

\subsection{Prompts for GPT-4 Evaluation}
\label{sec:prompts_gpt4_eval}
We show the evaluation prompts for \textit{GPT-4 + caption} and \textit{GPT-4 + detection} in Figure~\ref{fig:gpt4_caption} and Figure~\ref{fig:gpt4_detection}, respectively.

\begin{figure*}[t]
\footnotesize
\begin{AIbox}{Prompts for \textit{GPT-4 + Caption} Evaluation}
{\bf Prompt:} \\
{\footnotesize
\#\#\# You are an excellent evaluator. \\
\#\#\# Your assignment involves providing evaluations for given responses. \\
\#\#\# Each evaluation consists of *a caption*, *a question*, a *question type*, and *two corresponding answers*. Your task is to discern which answer is superior based on the **quality** and its alignment w.r.t the caption. \\
\#\#\# There are only two situations where you may choose 'unable to decide': \\
\#\#\#\# Situation one: The question type is 'close-ended' and both answers are correct or wrong.  \\
\#\#\#\# Situation two: Both answers contain considerable factual errors or ethical issues. \\
\#\#\# Otherwise, you should always choose a better answer by responding 'Answer1' or 'Answer2'. \\
\textasciitilde\textasciitilde\textasciitilde Caption \\
\{caption\} \\
\textasciitilde\textasciitilde\textasciitilde  \\
\textasciitilde\textasciitilde\textasciitilde Question \\
\{question\} \\
\textasciitilde\textasciitilde\textasciitilde  \\
\textasciitilde\textasciitilde\textasciitilde Answer1 \\
\{answer1\} \\
\textasciitilde\textasciitilde\textasciitilde  \\
\textasciitilde\textasciitilde\textasciitilde Answer2 \\
\{answer2\} \\
\textasciitilde\textasciitilde\textasciitilde  \\
\#\#\# You should ONLY output your vote 'Answer1', 'Answer2', 'unable to decide: situation one', or 'unable to decide: situation two' in the last line.
}
\end{AIbox}
\caption{\label{fig:gpt4_caption} Prompts for \textit{GPT-4 + Caption} Evaluation}
\end{figure*}

\begin{figure*}[htbp]
\footnotesize
\begin{AIbox}{Prompts for \textit{GPT-4 + Detection} Evaluation}
{\bf Prompt:} \\
{\footnotesize
\#\#\# You are an excellent evaluator. \\
\#\#\# Your assignment involves providing evaluations for given responses. \\
\#\#\# Each question will consist of a *list of objects* about an image, a *question* and *two corresponding answers*. Your task is to discern which response is superior based on the **quality of the answer** and its alignment w.r.t the objects.  \\
\#\#\# Each object in the object list will contains three keys, "bbox", "conf", "label" \\
\#\#\#\# "bbox" is a list of four numbers, which are the coordinates of the bounding box of the object in the image, the order is [x1, y1, x2, y2], where (x1, y1) is the top left corner of the bounding box, (x2, y2) is the bottom right corner of the bounding box. \\
\#\#\#\# "conf" is a number, which is the confidence of the object detection model. \\
\#\#\#\# "label" is a string, which is the label of the object. \\
\#\#\# There are only two situations where you may choose 'unable to decide': \\
\#\#\#\# Situation one: The question type is 'close-ended' and both answers are correct or wrong.  \\
\#\#\#\# Situation two: Both answers contain considerable factual errors or ethical issues. \\
\#\#\# Otherwise, you should always choose a better answer by responding 'Answer1' or 'Answer2'. \\

\textasciitilde\textasciitilde\textasciitilde Object List \\
\{object\_list\} \\
\textasciitilde\textasciitilde\textasciitilde  \\
\textasciitilde\textasciitilde\textasciitilde Question \\
\{question\} \\
\textasciitilde\textasciitilde\textasciitilde  \\
\textasciitilde\textasciitilde\textasciitilde Answer1 \\
\{answer1\} \\
\textasciitilde\textasciitilde\textasciitilde  \\
\textasciitilde\textasciitilde\textasciitilde Answer2 \\
\{answer2\} \\
\textasciitilde\textasciitilde\textasciitilde  \\

\#\#\# You should ONLY output your vote 'Answer1', 'Answer2', 'unable to decide: situation one', or 'unable to decide: situation two' in the last line.
}
\end{AIbox}
\caption{ \label{fig:gpt4_detection}Prompts for \textit{GPT-4 + Detection} Evaluation}
\end{figure*}

\subsection{Comparison of Alignment between Human and Different Methods}
Table~\ref{tab:baseline_and_human_alignment_with_num} shows the comparison of alignment between different evaluation methods and human evaluation. The number of evaluations for each settings are listed below the ratio.

\begin{table*}[htbp]
\small
\setlength{\tabcolsep}{3pt} 
\centering
\caption{Comparison of alignment between different evaluation methods and human evaluation, the aggregated column aggregates the results of 4 human evaluators (includes the number of valid evaluations in each setting). The agreement is calculated through a matching percentage.}
\vspace{1em}
\begin{tabular}{lccccll}
\toprule
\multirow{2}{*}{} 
 & \textbf{Evaluator 1} & \textbf{Evaluator 2} & \textbf{Evaluator 3} & \textbf{Evaluator 4}  & \textbf{Aggregated}\\
 & (Order 1) & (Order 1) & (Order 2) & (Order 2) & \\
  
 \midrule
\multirow{2}{*}{GPT-4 + detection w/o  Criteria} & 79.1\% & 78.2\% & 71.54\% & 72.48\% & 75.37\% \\ 
 & \textcolor{gray!50}{(106/134)} & \textcolor{gray!50}{(104/133)} & \textcolor{gray!50}{(93/130)} & \textcolor{gray!50}{(79/109)} & \textcolor{gray!50}{(101/134)} \\
\multirow{2}{*}{GPT-4 + detection w/  Criteria} & 82.71\% & 78.79\% & 72.87\% & 80.18\% & 81.2\% \\ 
 & \textcolor{gray!50}{(110/133)} & \textcolor{gray!50}{(104/132)} & \textcolor{gray!50}{(94/129)} & \textcolor{gray!50}{(89/111)} & \textcolor{gray!50}{(108/133)} \\ \midrule
\multirow{2}{*}{GPT-4 + caption w/o  Criteria} & 80.0\% & 81.72\% & 74.59\% & 82.14\% & 81.15\% \\
 & \textcolor{gray!50}{(152/190)} & \textcolor{gray!50}{(152/186)} & \textcolor{gray!50}{(135/181)} & \textcolor{gray!50}{(138/168)} & \textcolor{gray!50}{(155/191)} \\
\multirow{2}{*}{GPT-4 + caption w/ Criteria} & 78.72\% & 80.32\% & 74.58\% & 80.98\% & 80.53\% \\ 
 & \textcolor{gray!50}{(148/188)} & \textcolor{gray!50}{(151/188)} & \textcolor{gray!50}{(132/177)} & \textcolor{gray!50}{(132/163)} & \textcolor{gray!50}{(153/190)} \\ \midrule
 \multirow{2}{*}{GPT-4 + detection + caption w/o  Criteria} & 81.12\% & 83.42\% & 75.27\% & 84.12\% & 83.0\% \\ 
 & \textcolor{gray!50}{(159/196)} & \textcolor{gray!50}{(161/193)} & \textcolor{gray!50}{(140/186)} & \textcolor{gray!50}{(143/170)} & \textcolor{gray!50}{(166/200)} \\
 \multirow{2}{*}{GPT-4 + detection + caption w/ Criteria} & 80.41\% & 82.29\% & 74.86\% & 83.54\% & 83.16\% \\ 
 & \textcolor{gray!50}{(156/194)} & \textcolor{gray!50}{(158/192)} & \textcolor{gray!50}{(137/183)} & \textcolor{gray!50}{(137/164)} & \textcolor{gray!50}{(163/196)} \\ \midrule
\multirow{2}{*}{GPT-4V w/o  Criteria} & 81.62\% & 82.22\% & 75.42\% & 82.05\% & 82.8\% \\ 
 & \textcolor{gray!50}{(151/185)} & \textcolor{gray!50}{(148/180)} & \textcolor{gray!50}{(135/179)} & \textcolor{gray!50}{(128/156)} & \textcolor{gray!50}{(154/186)} \\
\multirow{2}{*}{GPT-4V w/  Criteria} & \textbf{78.61\%} & \textbf{80.30\%} & \textbf{73.00\%} & \textbf{77.84\%} & \textbf{88.02\%} \\ 
 & \textcolor{gray!50}{(158/201)} & \textcolor{gray!50}{(159/198)} & \textcolor{gray!50}{(146/200)} & \textcolor{gray!50}{(144/185)} & \textcolor{gray!50}{(147/167)} \\ \hline

\end{tabular}

\label{tab:baseline_and_human_alignment_with_num}
\end{table*}

\subsection{Evaluation Results of Claude-3-Opus as the Judge}
\label{app:claude_res}
The evaluation result using Claude-3-Opus as the judge is shown in Table~\ref{tab:claude_vote_llava}.

\begin{table*}[htbp]
\footnotesize
\centering
\caption{Number of wins/ties/loses for each model on each level, adopting \textbf{LLaVA-v1.5-13B} as the anchor and \textbf{Claude-3-Opus} as the judge. Models are sorted by overall win rates in descending order.
}
\vspace{1em}
\resizebox{\textwidth}{!}{
\begin{tabular}{@{}lccccccc@{}}
\toprule
\textbf{Models} & \textbf{Perception} & \textbf{Understanding} & \textbf{Applying} & \textbf{Analyzing} & \textbf{Evaluation} & \textbf{Creation} & \textbf{Win Rates } \\ 
\midrule

Qwen-VL-Chat        & 31/21/18            & 37/24/49               & 25/20/15          & 35/25/40           & 9/9/22             & 11/11/18          & 0.35               \\
LVIS                & 26/17/27            & 30/35/45               & 15/26/19          & 34/28/38           & 11/14/15           & 11/12/17          & 0.30               \\
mPLUG-Owl2          & 18/18/34            & 28/24/58               & 13/24/23          & 19/27/54           & 11/8/21            & 8/9/23            & 0.23               \\
MiniGPT-v2          & 20/16/34            & 23/28/59               & 11/23/26          & 23/22/55           & 11/6/23            & 4/9/27            & 0.22               \\
SEED-LLaMA          & 17/12/41            & 10/21/79               & 13/17/30          & 11/20/69           & 7/4/29             & 1/7/32            & 0.14               \\
InstructBLIP        & 12/18/40            & 13/14/83               & 10/11/39          & 11/22/67           & 6/7/27             & 2/7/31            & 0.13               \\
Fuyu-8B             & 9/16/45             & 4/20/86                & 5/11/44           & 6/14/80            & 2/1/37             & 0/1/39            & 0.06               \\
kosmos2             & 5/19/46             & 3/11/96                & 2/13/45           & 2/23/75            & 2/1/37             & 0/5/35            & 0.03               \\
BLIP2               & 4/11/55             & 2/11/97                & 4/9/47            & 1/11/88            & 1/2/37             & 0/1/39            & 0.03               \\ 

\bottomrule
\end{tabular}}

\label{tab:claude_vote_llava}
\end{table*}

\begin{table*}[htbp]
\centering
\setlength{\tabcolsep}{3pt}
\caption{Pairwise Agreement among Human Evaluators.}
\vspace{1em}
\begin{tabular}{lcccc}
\toprule
          & \textbf{Evaluator 1} & \textbf{Evaluator 2} & \textbf{Evaluator 3} & \textbf{Evaluator 4}  \\ \midrule
\multirow{2}{*}{Evaluator 1}     & -  & 0.88      & 0.80   & 0.82    \\
 & \textcolor{gray!50}{-} 
 & \textcolor{gray!50}{(198/225)} 
  & \textcolor{gray!50}{(177/220)}
 & \textcolor{gray!50}{(160/196)}\\
 
\multirow{2}{*}{Evaluator 2}     & 0.88  & -     & 0.85   & 0.83    \\
  & \textcolor{gray!50}{(198/225)} 
  & \textcolor{gray!50}{-}
  & \textcolor{gray!50}{(181/212)}
 & \textcolor{gray!50}{(161/194)}    \\
 
 \multirow{2}{*}{Evaluator 3}     & 0.80  &   0.85  & -   & 0.81    \\
 & \textcolor{gray!50}{(177/220)}
   & \textcolor{gray!50}{(181/212)}
  & \textcolor{gray!50}{-}

 & \textcolor{gray!50}{(153/189)} \\

  \multirow{2}{*}{Evaluator 4}     & 0.82  & 0.83     & 0.81  & -    \\
  & \textcolor{gray!50}{(160/196)}
  & \textcolor{gray!50}{(161/194)}
  & \textcolor{gray!50}{(153/189)}
  & \textcolor{gray!50}{-}

 \\ \bottomrule
\end{tabular}
\label{tab:human_correlation}
\end{table*}

\section{Discussions and Limitations}
\label{sec:limitations}

While MLLM-Bench strives to assess multimodal large language models (MLLMs) comprehensively, it cannot encapsulate the full diversity of real-world multimodal interactions, acknowledging the challenge of simulating the unpredictable variety of real-life tasks. 

\paragraph{Societal Impact} By leveraging per-sample criteria and advanced MLLMs as evaluators, our benchmark provides a more nuanced and accurate assessment of multimodal AI systems, which can lead to the development of more reliable and user-friendly AI applications, enhancing user experience and trust in AI technologies. Also, one needs to recognize potential negative impacts. By promoting automated evaluation methods, there is a risk that human judgment may be undervalued or overlooked. While our benchmark aims to align closely with human evaluations, it is crucial to maintain a balance and ensure that human oversight remains integral to the evaluation process.

\paragraph{Potential Subjectivity} The design of human-annotated per-sample criteria, which seeks to mirror human user experience, may introduce subjectivity, potentially affecting the consistency and generalizability of results. However, the primary objective of introducing per-sample criteria is to align model performance with human needs, as the benchmark is designed to evaluate the real-world capabilities of the models. While we acknowledge the existence of subjectivity in per-sample criteria, we posit it as a "positive" subjectivity, as it brings the model's performance closer to human expectations and practical needs.


\paragraph{Reproducibility} 
\label{sec:reproducibility}
To promote reproducibility, we have provided all the necessary code to replicate the results presented in this paper, along with the evaluation prompts detailed in Appendix~\ref{sec:4V_detail}. Reproducibility is a cornerstone of this research, underscoring its importance and our commitment to transparency and scientific rigor.


\paragraph{Positional Biases} The mention of length and position biases points to intrinsic limitations in GPT-4V's processing, see details in Appendix~\ref{sec:postional_bias}. These biases can affect the model's performance on the benchmark, potentially skewing results based on the length of input or the position of relevant information. This suggests a need for criteria that account for these biases, ensuring that the evaluation reflects the model's ability to understand and generate content impartially, regardless of these factors. As mentioned in \S~\ref{sec:evaluation_setting}, shuffling order of model-generated responses could migrate the position bias, as done by \citep{wang2023large,chen2023phoenix}.

\paragraph{Extensibility to Larger Scale Benchmark Dataset} One might be concerned that the current dataset size is not large enough. In this paper, The selection of 420 samples was a deliberate methodological choice, aimed at demonstrating the efficacy of our proposed evaluation paradigm (i.e. per-sample criteria for evaluation) for multimodal large language models (MLLMs) in open-ended tasks. Moreover, our approach is scalable. Moreover, we will continuously update and scale up our benchmark in the future version. 

\paragraph{Extensibility to More Recent MLLMs}
Within the scope of our continuous endeavors, we remain dedicated to the inclusion of the most recent models into our evaluative framework. To support this objective, an online leaderboard has been established, which openly encourages submissions from the global community. This platform is designed to facilitate the perpetual evaluation and juxtaposition of novel models against pre-established benchmarks, thereby promoting an atmosphere of relentless innovation and enhancement. Notably, to prevent evaluation leakage and deter potential benchmark manipulation, the per-sample criteria employed by this online leaderboard are confidential, accessible exclusively to the individual submitters.

The qualitative nature of benchmarks, especially in creative or ethical scenarios, also complicates the evaluation process.  Ethical considerations, despite being integrated into the framework, cannot capture the full spectrum of societal implications, with the fluidity of AI ethics demanding continuous updates to the benchmark. Acknowledging these limitations is vital for the nuanced application and interpretation of MLLM-Bench results, and underscores the necessity for iterative refinement to enhance the tool's relevance and evaluative accuracy.

\end{document}